\documentclass[lettersize,journal]{IEEEtran}
\usepackage{amsmath,amsfonts}
\usepackage{mathrsfs}
\usepackage{cite}
\usepackage{algorithmic}
\usepackage{array}
\usepackage[caption=false,font=normalsize,labelfont=sf,textfont=sf]{subfig}
\usepackage{textcomp}
\usepackage{stfloats}
\usepackage{url}
\usepackage{verbatim}
\usepackage{graphicx}
\usepackage{multirow}
\usepackage{booktabs}
\usepackage{makecell}
\usepackage{amssymb}
\usepackage{threeparttable}
\usepackage{xcolor}
\def\BibTeX{{\rm B\kern-.05em{\sc i\kern-.025em b}\kern-.08em
    T\kern-.1667em\lower.7ex\hbox{E}\kern-.125emX}}
\usepackage{balance}
\begin{document}
\title{VidFormer: A novel end-to-end framework fused by 3DCNN and Transformer for Video-based Remote
	Physiological Measurement}
\author{Jiachen Li, Shisheng Guo, \IEEEmembership{Member, IEEE}, Longzhen Tang, Guolong Cui, \IEEEmembership{Senior Member, IEEE},  Lingjiang Kong, \IEEEmembership{Senior Member, IEEE} and Xiaobo Yang
\thanks{This work was supported in part by the National Natural Science Foundation of China under Grant 62371110, in part by the Municipal Government of Quzhou under Grant 2022D008 and Grant 2022D005, and in part by the 111 Project under Grant B17008. \emph{(Corresponding author: Shisheng Guo.)}}
\thanks{The authors are with the School of Information and Communication Engineering, University of Electronic Science and Technology of China, Chengdu 611731 China.}
\thanks{Shisheng Guo and Guolong Cui are also with the Yangtze Delta Region Institute, University of Electronic Science and Technology of China, Quzhou 324000, China (e-mail: ssguo@uestc.edu.cn).}}

\markboth{Journal of \LaTeX\ Class Files,~Vol.~18, No.~9, September~2020}%
{How to Use the IEEEtran \LaTeX \ Templates}

\maketitle

\begin{abstract}
Remote physiological signal measurement based on facial videos, also known as remote photoplethysmography (rPPG), involves predicting changes in facial vascular blood flow from facial videos. While most deep learning-based methods have achieved good results, they often struggle to balance performance across small and large-scale datasets due to the inherent limitations of convolutional neural networks (CNNs) and Transformer. In this paper, we introduce VidFormer, a novel end-to-end framework that integrates 3-Dimension Convolutional Neural Network (3DCNN) and Transformer models for rPPG tasks. Initially, we conduct an analysis of the traditional skin reflection model and subsequently introduce an enhanced model for the reconstruction of rPPG signals. Based on this improved model, VidFormer utilizes 3DCNN and Transformer to extract local and global features from input data, respectively. To enhance the spatiotemporal feature extraction capabilities of VidFormer, we incorporate temporal-spatial attention mechanisms tailored for both 3DCNN and Transformer. Additionally, we design a module to facilitate information exchange and fusion between the 3DCNN and Transformer. Our evaluation on five publicly available datasets demonstrates that VidFormer outperforms current state-of-the-art (SOTA) methods. Finally, we discuss the essential roles of each VidFormer module and examine the effects of ethnicity, makeup, and exercise on its performance.

\end{abstract}


\section{Introduction}
\IEEEPARstart{P}{hysiological} signals measurement has long been a significant research domain that can provide insights into the health condition of the human body\cite{faust2018deep}. Physiological signals such as heart rate (HR), heart rate variability (HRV), and respiratory rate (RR) are particularly crucial for human body health accurate assessment\cite{rajendra2006heart,kleiger2005heart}. In recent years, the Remote Photoplethysmography (rPPG) has gained considerable attention as a research focal point\cite{choi2005blind,de2013robust,de2014improved,yan2018contact,niu2018synrhythm,yu2019remote,yue2023facial}. In comparison to traditional contact-based methods like Electrocardiography (ECG) \cite{nishime2000heart} and Photoplethysmography (PPG)\cite{temko2017accurate}, rPPG offers distinct advantages by eliminating the physical contact with patients. Consequently, it mitigates concerns related to allergies, disease transmission, and patient discomfort\cite{niu2019rhythmnet}. Therefore, rPPG is widely used in fields such as emotion computing, intelligent assistants, and biometric recognition\cite{yu2021facial,de2013robust}.

The rPPG approaches utilize RGB cameras for the purpose of capturing facial skin and subsequently analyzing it in order to obtain PPG signals\cite{lee2020meta,lu2021dual,chen2018video}. This principle of rPPG is that the absorption rates of hemoglobin, oxyhemoglobin, and melanin in the blood of human skin exhibit variations across different spectra, and their concentrations fluctuate in response to the movements of  heart\cite{zonios2006modeling}. Specifically, during cardiac contraction, the blood volume within the blood vessels increases, causing an escalation in the absorption of specific light spectra. Conversely, when the cardiac diastole, the blood volume within the vessels diminishes, resulting in a reduction in the absorption of light within that particular spectrum\cite{verkruysse2008remote,wang2016algorithmic,de2013robust}. Accordingly, through the analysis of facial skin features observed in the video, the PPG signal of the human body can be obtained from facial video. Through the analysis of remote Blood Volume Pulse (BVP), the physiological parameters including HR, HRV, and RR of human body can be feasible to measure and evaluate. Nevertheless, BVP are susceptible to interference from external factors such as variations in ambient lighting, head movements, obstructions on the face, and different facial expressions, thereby generating non-periodic noise signals\cite{cheng2021deep,lokendra2022and}. To overcome these challenges, the traditional approaches applies blind source separation techniques and skin reflection models to evaluate alterations in the illumination of human skin, thus facilitating the extraction of BVP\cite{Poh:10,lewandowska2011measuring}. Nonetheless, these methods heavily depend on prior knowledge, e.g., De Hann \textit{et al.} assumes that under white light, people with different skin tones have the same color standard\cite{de2013robust}, hindering their application in diverse environmental conditions.   

With the progressive advancement of deep learning, an increasing number of methodologies employing deep neural networks (DNNs) for acquiring BVP have been proposed. These methodologies accomplish the mapping of blood volume fluctuations in facial video skin to BVP through the training and fitting of models. The most extensively trained neural network type is the convolutional neural network (CNN). Initially, Chen \textit{et al.} introduced the utilization of 2DCNN for processing rPPG signal tasks\cite{chen2018deepphys}, subsequently leading to the development of numerous other two-dimensional neural networks\cite{hu2021robust,yu2019remote,gideon2021way}. Nonetheless, the strong temporal correlation between BVP and facial video data makes it arduous to model long-term time series exclusively employing two-dimensional convolution\cite{botina2022rtrppg}. Thus, Yu \textit{et al.} proposed the adoption of 3D Convolution Neural Network (3DCNN)\cite{yu2019remote}, and Yu \textit{et al.} recommended the use of Transformer network architectures\cite{yu2021transrppg}. 3D convolution operations enable effective modeling of video data, facilitating the acquisition of temporal and spatial information\cite{zhu2018continuous}, while Transformers inherently possess the capability of long-term global modeling\cite{vaswani2017attention}. However, convolution operations lack effective global modeling capabilities, and Transformers lack a corresponding inductive bias and local perception abilities\cite{liu2024transformer}. Consequently, effectively addressing rPPG tasks by overcoming the deficiencies of both approaches is a considerably intriguing issue.

In this paper, we proposed a novel fusion framework that combines CNN with Transformer named VidFormer. Our framework successfully combines the Transformer's exceptional global modeling capability and the CNN's inductive bias, facilitating multi-level information exchange and fusion. As a result, it effectively accomplishes the reconstruction task of BVP signal. The proposed framework comprises of five modules: the Stem module, Local Convolution Branch, Global Transformer Branch, CNN and Transformer Interaction module (CTIM), and rPPG generation module (RGM). The Stem module serves to localize and eliminate redundant video information while preserving the primary video information. Subsequently, the simplified video information is fed into the CTIM to extract relevant information. Owing to the non-uniform blood distribution in the human face and the impact of effective scattering area and temporal variations of facial blood vessels, we utilized the CNN branch in our proposed CTIM method to capture local information of the human face. Additionally, we developed a spatial-temporal multi-head attention module for the CNN branch, enabling us to effectively capture the temporal variation patterns exhibited by the human face. Moreover, we employed the Transformer branch in CTIM to extract global information encompassing facial movements, changes in ambient lighting, and other relevant factors. Furthermore, we devised a multi-head attention mechanism specifically tailored to the characteristics of video data. In the RGM module, we processed the feature information obtained from CTIM to generate an rPPG signal, which integrates the spatial-temporal information, global and local information of the input data.

Overall, our proposed CNN Transformer fusion framework has the following key contributions:
\begin{itemize}
	\item We have proposed a novel end-to-end framework to leverage the complementary strengths of 3DCNN and Transformer namely VidFormer, thereby enhancing the modeling capabilities for rPPG. By fusing the inductive bias ability inherent in CNN with the global modeling prowess of Transformer, we achieve an effective integration. Moreover, our framework effectively tackles the task of rPPG from two distinct perspectives, namely, global modeling and local induction.
	\item We propose global attention mechanisms for 3DCNN that consider both the temporal and spatial dimensions of the input data e.g., Global Attention 3DCNN (GA-3DCNN). This design ensures that the convolution module focus on the extraction of spatial-temporal global features, thereby augmenting the global modeling capacity of the convolution branch.
	\item Given that traditional Transformer typically focus on the characteristics of input data from a single dimension, it becomes challenging in effectively capturing the multi-dimensional features of video data. To address this limitation, we propose a refinement by segregating the multi-head attention mechanism in Transformers into distinct temporal and spatial attention mechanisms e.g, Spatial-Time Multi-headed Self-attention (ST-MHSA). This makes Transformer concentrate on the spatial-temporal characteristics of features individually.
	\item To facilitate the integration of Transformer and CNN features, we devised the CTIM to enable efficient information exchange between the two architectures. This module effectively facilitates the communication of both global and local information, thereby enriching the informational exchange within each branch.
\end{itemize}

We conducted extensive experiments on multiple standard datasets, i.e., UBFC-rPPG\cite{bobbia2019unsupervised}, PURE\cite{stricker2014non}, DEAP\cite{koelstra2011deap}, ECG-fitness\cite{vspetlik2018visual}, and COHFACE\cite{heusch2017reproducible}. The experimental results show that our proposed method is superior to other state-of-the-art methods.

\section{Related Work}
The measurement techniques for estimating the BVP from facial videos encompass blind source separation, skin reflection model, and deep learning approaches. 

Blind source separation was one of the initial methods employed for rPPG measurement. This technique postulates that the facial skin tone variations arise as a linear combination of both the PPG signals and the ambient environmental signals. It assumes the independence of each signal source and estimates the coefficients for each source\cite{choi2005blind}. By calculating the confusion matrix, the signals can be successfully separated. Macwan \textit{et al.} utilized prior knowledge, namely auto-correlation constraints and skin chromaticity change constraints, to guide ICA in extracting BVP\cite{macwan2018remote}, while Favila \textit{et al.} utilized ICA for preprocessing to improve the accuracy of rPPG signal extraction\cite{favilla2018heart}.

The skin reflection model mainly utilizes the principle of skin's reflection of light and converts the skin chromaticity in the RGB color space to other color spaces, in order to obtain better rPPG estimation. Haan et al proposed that noise caused by motion can be eliminated by projecting skin chromaticity into different color spaces\cite{de2013robust}. Moreover, to enhance the efficacy of mitigating shadow noise and the influence of diverse skin tones resulting from motion, a technique based on standardizing skin tones under white-light conditions has been proposed\cite{wang2016algorithmic}.

With the development of DNNs, numerous rPPG detection methods based on DNNs have been proposed\cite{chen2018video,chen2018deepphys,hu2021robust,yu2019remote,gideon2021way}. Chen \textit{et al.} introduced DeepPyhs, a system based on the skin reflection motion model and the CAN-based appearance model. This system effectively separates and extracts physiological signals using different frequencies of various signals\cite{chen2018deepphys}. Additionally, in order to achieve continuous HR detection in challenging environments, Tang \textit{et al.} proposed the use of CNN to segment skin pixels and the combination of PCA and ICA for signal analysis\cite{tang2018non}. Niu \textit{et al.} argued that noise signals caused by facial movements and surrounding environments are highly coupled with physiological signals. To effectively isolate physiological signals, Niu \textit{et al.} proposed the generation of MSTmaps from videos and the development of cross-verified feature disentangling (CVD) to decouple physiological signals from noise signals\cite{niu2019rhythmnet}. Owing to the temporal context inherent in videos,  LSTM\cite{gao2021lstm,gao2022remote,botina2020long}, GRU\cite{botina2020long}, Transformer\cite{yu2021transrppg,liu2024rppg,zhang2023demodulation,park2022self,yu2022physformer}, and 3DCNN\cite{yu2019remote,botina2022rtrppg,kuang2023shuffle,yin2022pulsenet} have been proposed for capturing long time series information. The Transformer model, known for its strong parallel processing capabilities and global modeling power, is extensively utilized in modeling diverse long time series data. Yu \textit{et al.} introduced the application of Transformer in extracting BVP\cite{yu2021transrppg}. Yu and Liu \textit{et al.} utilized Vision Transformer (ViT) to construct mask autoencoders for self-supervised training, enabling the extraction of disregarded intrinsic self-similarity priors\cite{liu2024rppg}. Park \textit{et al.} used video vision Transformer (ViViT) to process RGB and NIR data, and designed appropriate Transformer structures from a video perspective to extract BVP\cite{park2022self}. Furthermore, 3DCNN is well-suited for video data processing due to its effective capacity in extracting three-dimensional information and the ability of Inductive bias\cite{guo2022cross}. Yu \textit{et al.} compared the processing results of two-dimensional convolution and three-dimensional convolution for rPPG tasks, and found that the results of three-dimensional convolution were much better than those of two-dimensional convolution\cite{yu2019remote}.  Kuang \textit{et al.} utilized the global context information of input data to construct a global attention mechanism for BVP signal reconstruction.\cite{kuang2023shuffle}. However, due to the huge demand for graphics memory in 3DCNN, it is difficult for 3DCNN to form a deeper network structure. Botina \textit{et al.} constructed a lightweight 3DCNN model for rPPG tasks\cite{botina2022rtrppg}. To date, no fusion framework has been proposed that integrates both 3DCNN and Transformer, specifically designed to address the challenges associated with rPPG tasks and mitigate the inherent limitations of these individual models\cite{xu2021vitae,cao2022random}. Therefore, we proposed a novel fusion framework that integrates the advatanges of 3DCNN and Transformer to extract BVP from both local and global features, effectively addressing the inherent limitations of these models in this paper.

\section{Method}
\subsection{Analysis of Skin Reflection Model} \label{Analysis}
The skin reflection model proposed in \cite{de2013robust,wang2016algorithmic} is based on the dichromatic model\cite{tominaga1994dichromatic}, which can be expressed as 
\begin{equation}
	\label{equation0}
	\begin{split}
		C_k(t)=I(t)(v_s(t)+v_d(t))+v_n(t)
	\end{split}
\end{equation}
where $C_k(t)$ denotes the RGB channels (in column) of the $k$-th skin-pixel. $I(t)$ denotes the luminance intensity level, which absorbs the intensity changes due to the light source as well as to the distance changes between light source, skin tissue and camera. $v_s(t)$ and $v_d(t)$ are the specular reflection and diffuse reflection. $v_n(t)$ indicates the quantization noise of the camera sensor. 

However, the model focuses more on displaying the results of light reflected and absorbed by the skin in the RGB image of the camera, which is not suitable for rPPG reconstruction. Actually, the rPPG signal reconstruction task involves modeling the mapping relationship between BVP signals and skin color. Assuming the ground truth BVP signal is $y_t$, then we have 
\begin{equation}
	\label{equation01}	
	C_k(t)=I(t)[\ell_s(y_t(t),\rho,t)+\ell_d(y_t(t),\rho,t)]+v_n(t)
\end{equation}
where $\rho$ denotes the location of a single skin pixel, $\ell_d$ and $\ell_s$ are the mapping of the specular reflection and diffuse reflection. We posit a robust correlation between variations in BVP signals and changes in skin color. This correlation arises from the fluctuating volume of blood within superficial blood vessels, which governs both specular and diffuse light reflections, consequently eliciting temporal shifts in skin color. $\rho$ elucidates that diverse lighting conditions across distinct skin regions yield disparate specular and diffuse reflection\cite{zonios2006modeling}.

\begin{figure}[h]
	\centering
	\includegraphics[scale=0.30]{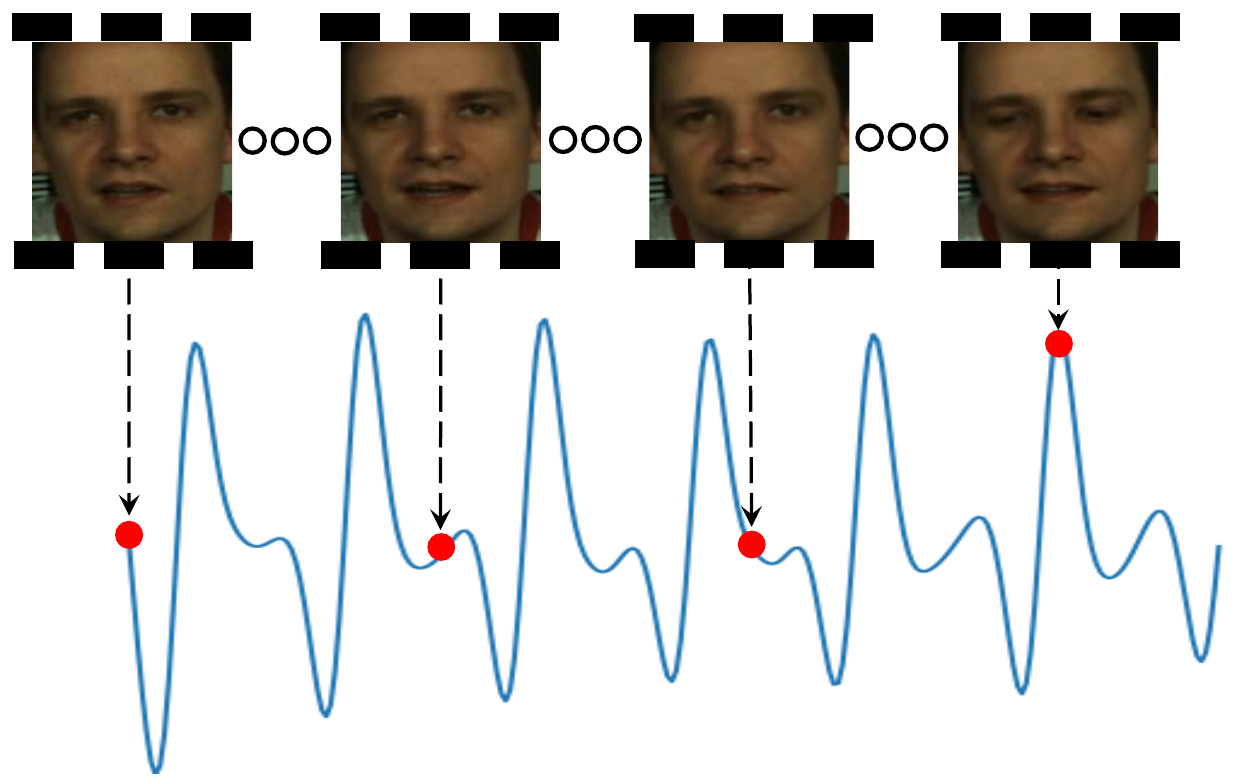}
	\caption{The potential mapping relationship between each frame in the video and the BVP signal.}
	\label{figure0}
\end{figure}

Actually, when the face is completely stationary, the illumination remains constant, and the video frame rate and BVP signal sampling rate align, the skin color and blood flow of the face are in a bijective relationship as shown in Fig. \ref{figure0}. 
However, the movement of the human head and changes in lighting can modulate this relationship, resulting in non-bijection. We believe in that the $\ell$ mapping function should include random functions related to human motion, changes in lighting, and other time factors. Therefore, $\ell$ is not only related to $\rho$ and $y_t(t)$, but also to $t$. 

Based on the above analysis, the influence of changes in blood flow within cutaneous blood vessels on the absorption spectra of the human skin results in discernible periodic alterations within the BVP. Simultaneously, the rPPG signal becomes intricately entwined with noise signals stemming from ambient lighting conditions and human head movements. These coupled signals are reflected in the RGB video data. Therefore, we utilize the prior knowledge mentioned above to design VidFormer and achieve accurate reconstruction of BVP signals.

\subsection{Model Overview}
\begin{figure*}[h]
	\centering
	\includegraphics[scale=0.42]{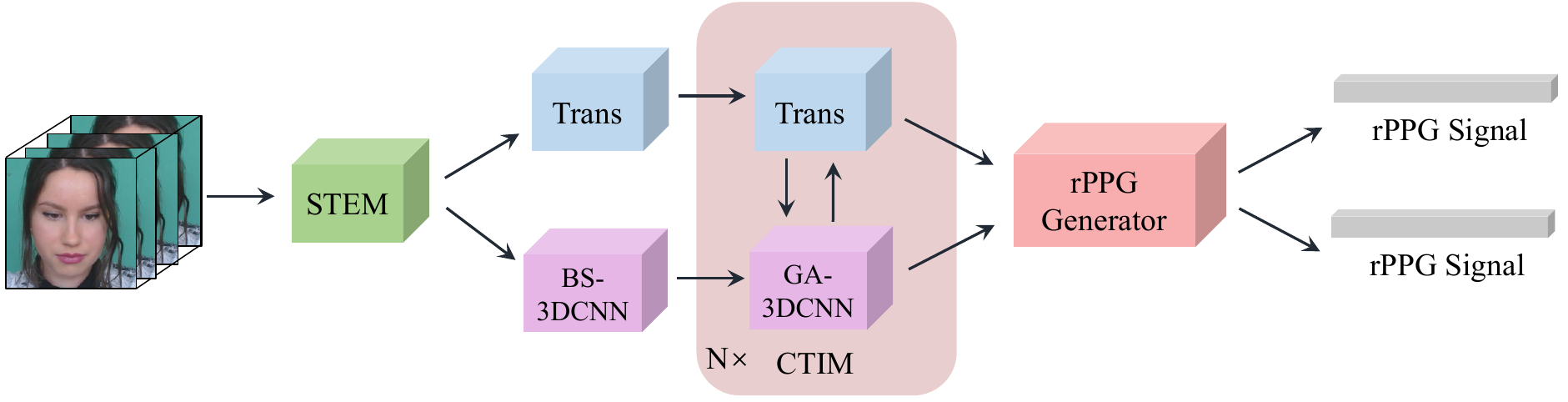}
	\caption{The overall framework of VidFormer. VidFormer leverages 3DCNN and Transformers to extract local and global features from input facial videos, respectively, and facilitates the interaction and fusion of these features. Additionally, VidFormer is designed with a modular structure that enables the efficient formation of a deeper network.}
	\label{figure1}
\end{figure*}

VidFormer employs a dual-branch architecture, comprising the Local Convolution Branch and Global Transformer Branch, to capture both local and global features as shown in Fig. \ref{figure1}. 
Firstly, the training data will be fed into a stem block consisting of a 3D convolution layer with a kernel size of 3 for preliminary feature extraction. Subsequently, the Local Convolution Branch is characterized by a hierarchical structure and a spatiotemporal convolutional attention mechanism. This branch employs convolutional operations to extract local information progressively, thus enlarging the receptive field. This approach directs the attention of  network towards to the skin regions where color changes most aptly align with the rPPG signal in time domain. Concurrently, the Transformer branch undertakes a comprehensive modeling of the representation of BVP within video data from a global perspective. By designing ST-MHSA, the emphasis of transformer is placed on delineating strong correlations among disparate image regions from a spatial standpoint and exploring the periodicity of signals in the time domain. This facilitates an effective modeling of the global influences of environmental lighting and surrounding ambient noise, thus it achieves decoupling between environmental noise and rPPG signal. CTIM promotes the fusion of locally and globally extracted information from the two branches, thereby synthesizing a comprehensive feature for rPPG signal. In this framework, the Local Convolution Branch refines its hierarchical features by incorporating the gleaned global information, thereby mitigating undue local influences on the branch. Simultaneously, the Transformer branch leverages the acquired local information to induce bias effectively, thereby enhancing the convergence speed of the Transformer branch. This sophisticated interplay within CTIM optimally integrates local and global features, contributing to the refined and expedited convergence of the overarching model.

\subsection{Local Convolution Branch}   
We design Local Convolution Branch for local feature extraction of data to facilitate the attention of network to key areas of input data. We expect these regions to better reflect the mapping relationship between BVP signals and skin color changes, as well as to be less affected by environmental noise. This branch consists of two parts: GA-3DCNN and BS-3DCNN.

\subsubsection{GA-3DCNN}
\begin{figure*}[h!]
	\centering
	\includegraphics[scale=0.42]{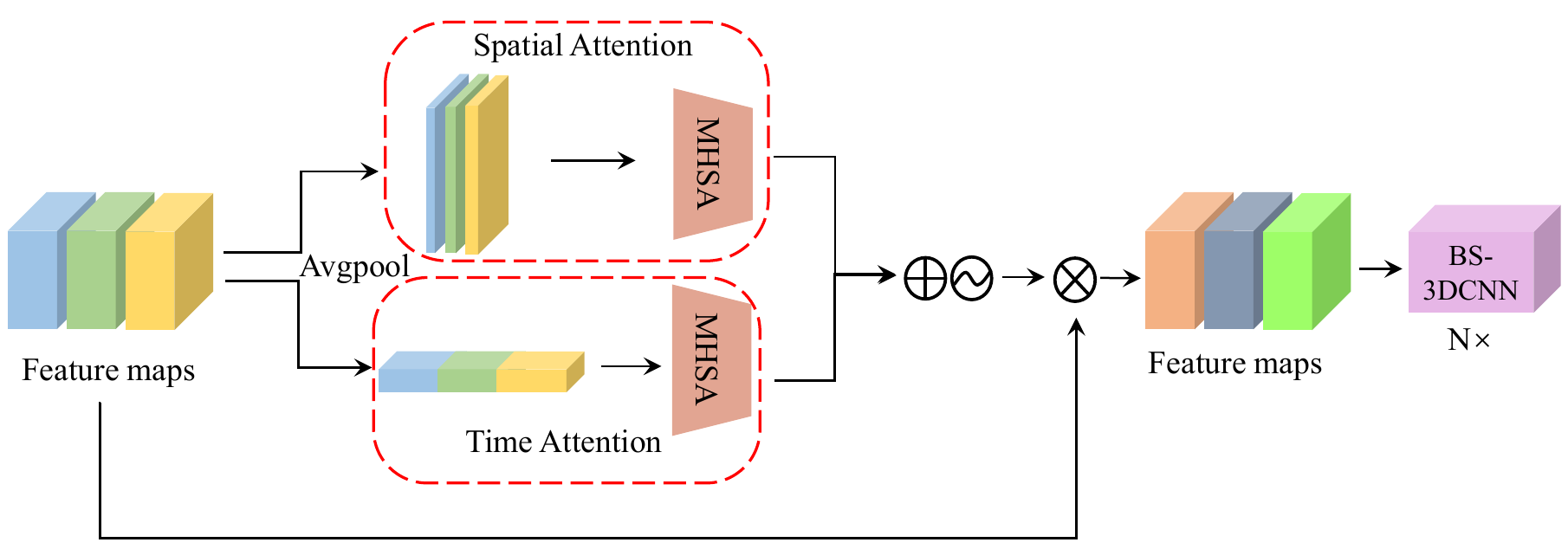}
	\caption{The illustration of the GA-3DCNN module. GA-3DCNN incorporates a global attention mechanism to assist BS-3DCNN in focusing on critical spatiotemporal regions within the input video.}
	\label{figure2}
\end{figure*}

GA-3DCNN is based on BS-3DCNN and designs temporal and spatial attention mechanisms to enhance the weights of input data as shown in Fig. \ref{figure2}.
The GA-3DCNN model incorporates Spatial Attention, Time Attention, and a BS-3DCNN component. The Spatial Attention focuses on the global spatial information across the entire image frame and accentuates correlations among different spatial patches. Consequently, this process encourages the BS-3DCNN to prioritize patches exhibiting pronounced correlations within each other when analyzing local information. Addtional, the Time Attention places greater attention on the temporal correlations in the input features of GA-3DCNN. It posits that signals exhibiting periodicity need to increase weighting of it to effectively capture features associated with periodic signals. This approach ensures that GA-3DCNN effectively captures both spatial and temporal dynamics changes, thereby enhancing its capacity to discern and model intricate patterns within the input data.

\begin{figure}[t!]
	\centering
	\includegraphics[scale=0.37]{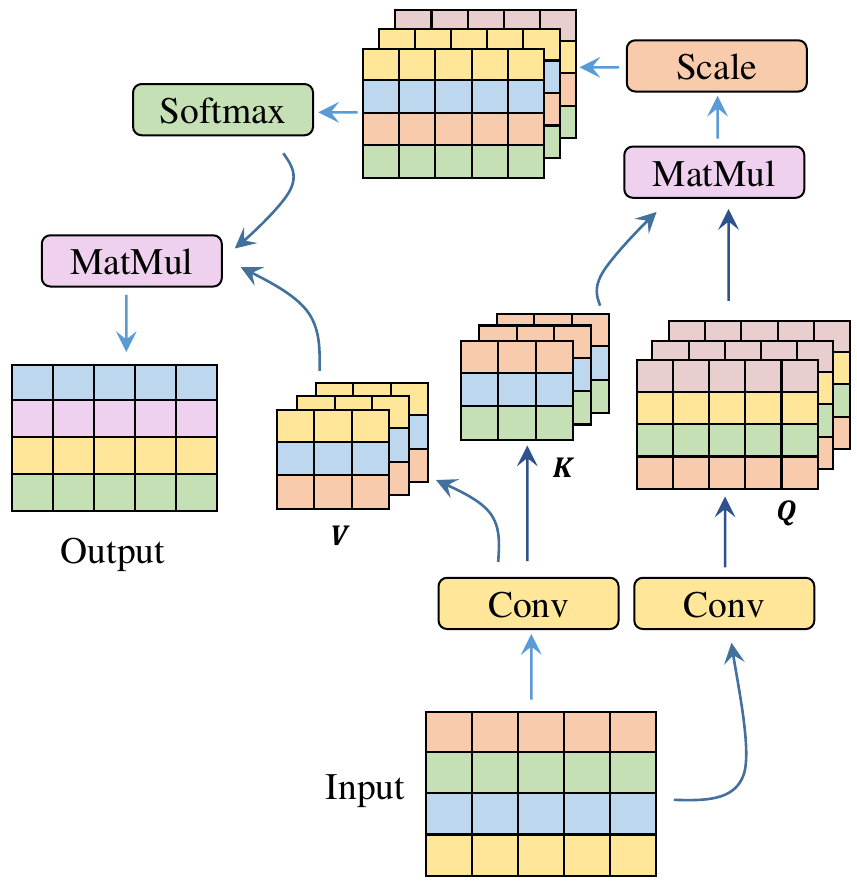}
	\caption{The schematic diagram of multi-head attention mechanism in Spatial Attention and Time Attention. }
	\label{figure3}
\end{figure}

Specifically, assuming that the input matrix of GA-3DCNN is $\mathbf{X}_{C}^{(k)}\in \mathbb{R}^{B\times C\times T\times H \times W}$, where $B$, $C$, $T$, $H$, $W$ represent the batch size, number of channels, frame length, height, and width of the input features respectively. $k$ is the input data of the $k$th layer. After undergoing different adaptive average pooling, $\mathbf{X}_{C}^{k-1}$ can be written as
\begin{equation}
	\label{equation1}
	\begin{split}
		&\mathbf{X}_s^{(k)}=\psi_{s} (\mathbf{X}_{C}^{(k-1)})\\
		&\mathbf{X}_t^{(k)}=\psi_{t} (\mathbf{X}_{C}^{(k-1)})
	\end{split}
\end{equation}
where $\psi_{s}$ and $\psi_{t}$ denote the adaptive avg-pool function. Additional, $\mathbf{X}_s^{(k)}\in\mathbb{R}^{B\times C \times 1 \times H \times W}$ and $\mathbf{X}_t^{(k)}\in\mathbb{R}^{B\times C\times T \times 1 \times 1}$ are the input data for Spatial Attention and Time Attention respectively and the form of multi-head attention mechanism is shown in the Fig. \ref{figure3}.
Therefore, the output after multi-head attention in Fig. \ref{figure3} can be written as
\begin{equation}
	\mathbf{X}_{o,i}^{(k)}=\xi(\frac{(\varsigma_1(\mathbf{X}_i^{(k)})^{T}\varsigma_2(\mathbf{X}_i^{(k)}))}{\sqrt{C}})\varsigma_1(\mathbf{X}_i^{(k)}), (i=s,t)
	\label{equation2}
\end{equation}
where $\varsigma$ denotes the convolution operation, $\xi$ is the Softmax function and $\mathbf{X}_{o,i}^{(k)}$ represents the output of Spatial-Attention or Time-Attention. As both Spatial Attention and Time Attention  excavate features from input data, the neural network undergoes a reconsideration of feature distributions, prompting a consequential realignment of model parameters. This phenomenon engenders an amplification in the variance of layer weights within the network and alleviate effect on the disappearance of gradients. Therefore, the weighted input obtained by 3DCNN can be represented as
\begin{equation}
	\label{equation3}
	\mathbf{X}_{g}^{(k)}=\sigma(\mathbf{X}_{o,s}^{(k)}+\mathbf{X}_{o,t}^{(k)})\circ \mathbf{X}_{C}^{(k-1)}
\end{equation}
where $\sigma$ is the sigmoid function and $\circ$ denotes the matrix dot product. Subsequently, the weighted features will be fed into BS-3DCNN for feature extraction.

\subsubsection{BS-3DCNN}
\begin{figure}[h]
	\centering
	\includegraphics[scale=0.36]{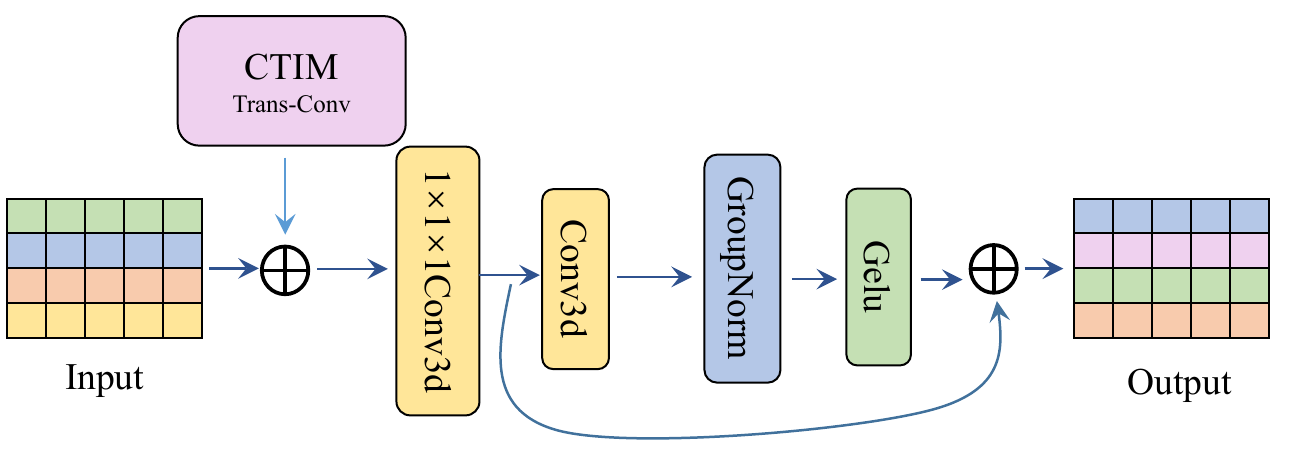}
	\caption{The illustration of BS-3DCNN. BS-3DCNN is designed for feature extraction from input data.}
	\label{figure4}
\end{figure}

The BS-3DCNN is employed for the purpose of extracting localized features from the input data. This is facilitated by its inherent hierarchical architecture, which spans the entirety of the 3DCNN branch. Consequently, the 3DCNN consistently iterates through the extraction of pertinent local features, thereby progressively enlarging its receptive field. This augmentation empowers the network to model the global information of input data and the BS-3DCNN framework can be seen in the Fig. \ref{figure4}.
 With the extraction of BS-3DCNN, the input from Spatial Attention and Time Attention $\mathbf{X}_{g}^{(k)}$ can be transformed into $\mathbf{X}_{C}^{k}$
 \begin{equation}
 	\label{equation4}
 	\begin{split}
 			\mathbf{X}_{C}^{(k)}=&Gelu(GroupNorm(\varsigma_1(\varsigma_2(\mathbf{X}_{tc}^{(k-1)}+\mathbf{X}_{g}^{(k)}))))\\
 			&+\varsigma_2(\mathbf{X}_{tc}^{(k-1)}+\mathbf{X}_{g}^{(k)})
 	\end{split}
 \end{equation}
 where $\mathbf{X}_{tc}^{k-1}$ indicates the output of the $(k-1)$th Trans-Conv Block in CTIM. $\varsigma_1$ and $\varsigma_2$ are the Convolution layer with $kernelsize=3$ and $kernelsize=1$ respectively. After the aforementioned architectural design, the BS-3DCNN adeptly captures localized feature information within the input feature. This extracted features encompasses both periodic signal components and environmental noise signals. Subsequently, Spatial Attention and Time Attention mechanisms are employed to weight the input feature information, leveraging a spatiotemporal perspective. This weighting compels a reorganization of input feature distributions along the weighted direction, thereby enhancing the capacity of network for information fitting. Nonetheless, convolutional operations primarily focus on amalgamating the local information of features. Despite the gradual expansion of the receptive field facilitated by pool function within hierarchical feature extraction structures, it is undeniable that certain features may undergo information loss during the pooling process. Hence, to address this concern, we introduce a Transformer branch.

\subsection{Global Transformer Branch}
\label{GTB}
As the commendable ability of transformer in global modeling, we devised a dedicated Transformer branch to extract comprehensive global information from the input data. We aim to devise a Global Transformer branch to effectively capture the nuanced dynamics cause by head movements and lighting variations. This is because during a short time window, the changes in lighting are very slow as indicated in Fig. \ref{equation01}.  
\begin{figure}[h!]
	\centering
	\includegraphics[scale=0.31]{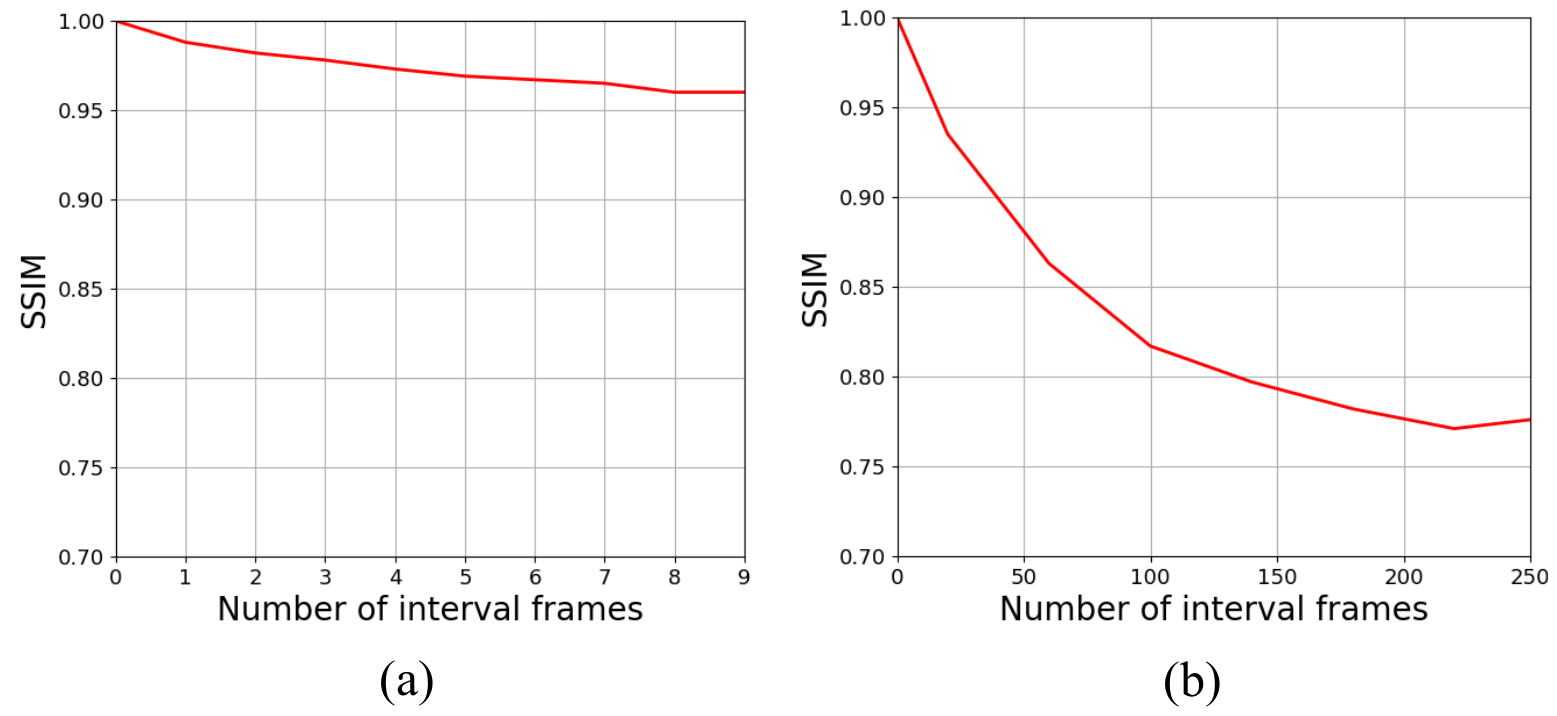}
	\caption{The relationship between SSIM and number of interval frames.}
	\label{figure01}
\end{figure}

Fig. \ref{equation01} demonstrates a gradual decline in Structural Similarity Index (SSIM)\cite{hore2010image} as the number of interval frames increases. However, it is noteworthy that within a brief timeframe, the SSIM value remains relatively stable, suggesting minimal variations in both environmental lighting and skin color over short durations.

Although Local Convolution Branch is structured hierarchically and incorporates GA-3DCNN, due to the structural limitations of small convolution kernels, using Local Convolution Branch alone requires a sufficiently deep network to expand the receptive field and be sensitive to changes in lighting, personnel movement, and so on. However, deeper network architectures are susceptible to challenges such as unstable training and convergence difficulties. To address these concerns, we integrate Transformer architecture to facilitate comprehensive global information extraction.

\begin{figure}[h!]
	\centering
	\includegraphics[scale=0.27]{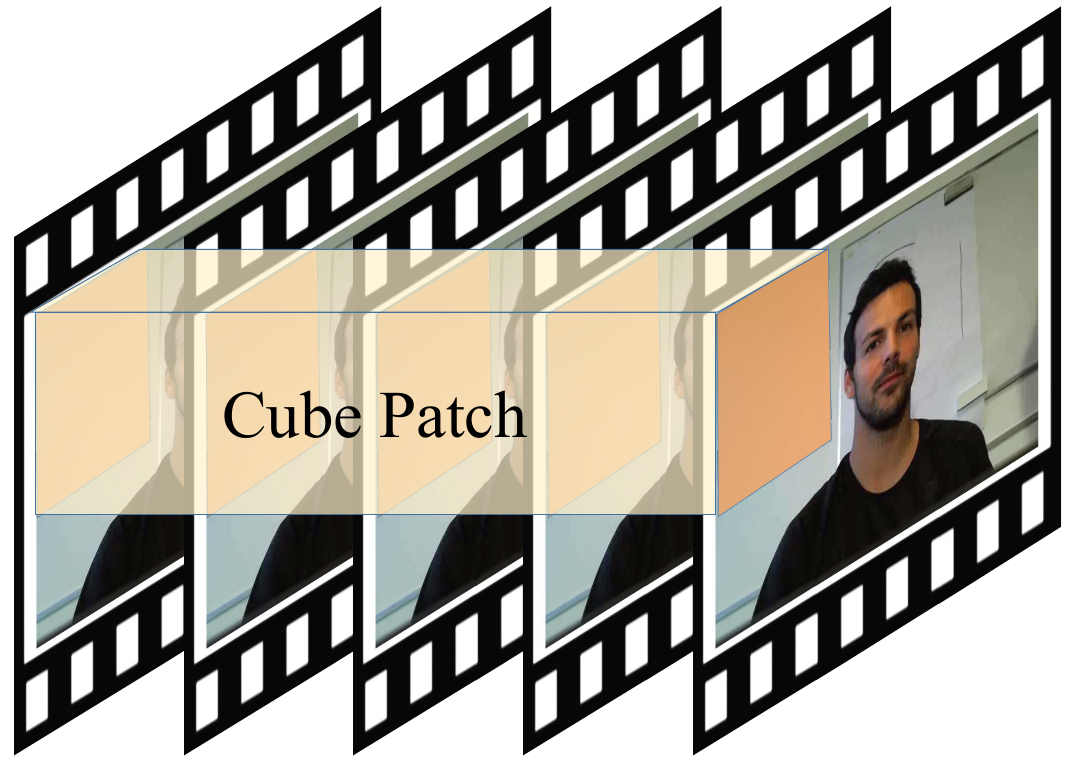}
	\caption{The process of cube patch of video data, where $h$ and $t$ denote the hight and the frame of input video.}
	\label{figure5}
\end{figure}

\begin{figure*}[h!]
	\centering
	\includegraphics[scale=0.45]{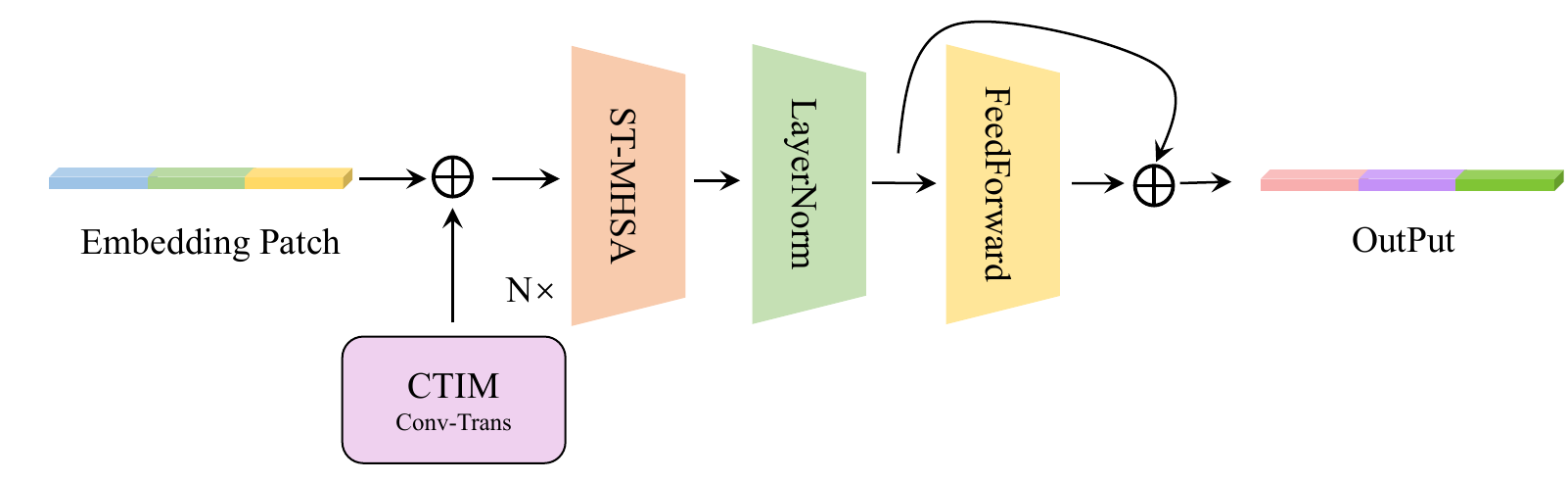}
	\caption{The illustration of Global Transformer Branch. The Global Transformer Branch based on the Transformer architecture is designed to extract global information from the input data.}
	\label{figure7}
\end{figure*}

Given the format of input data as video data, we undertook a redesign of the multi-head attention mechanism as ST-MHSA, tailored specifically for the extraction of intricate video information features. 
Firstly, based on the ViViT\cite{arnab2021vivit}, we slice the video data into cube patches to ensure that each ptach contains both temporal and spatial information as show in Fig. \ref{figure5}.
The obtained cube slices are embedded and add the positional information for each cube slice, then the cube slices are input into the Transformer branch. This process can be expressed in mathematical form as
\begin{equation}
	\label{equation5}
	\mathbf{X}^{(0)}_{T}=Lin(\aleph(I))+Pos
\end{equation}
where $\mathbf{X}^{(0)}_{T}\in \mathbb{R}^{B\times P \times D}$ denotes the input of the Global Transformer Branch, $P$ is the number of patches and $D$ indicates the embedding dim. $Lin$ is the linear layer. $I\in \mathbb{R}^{B\times C\times T\times H\times W}$ represents the input of the VidFormer and $\aleph$ is the cube slice operation. In addition, $Pos$ is a random number that satisfies a Gaussian distribution with a mean of 0 and a variance of 1. Subsequently, $\mathbf{X}^{0}_{T}$ is inputted into the Global Transformer Branch as shown in Fig. \ref{figure7}.

\subsubsection{ST-MHSA}
\begin{figure}[h]
	\centering
	\includegraphics[scale=0.34]{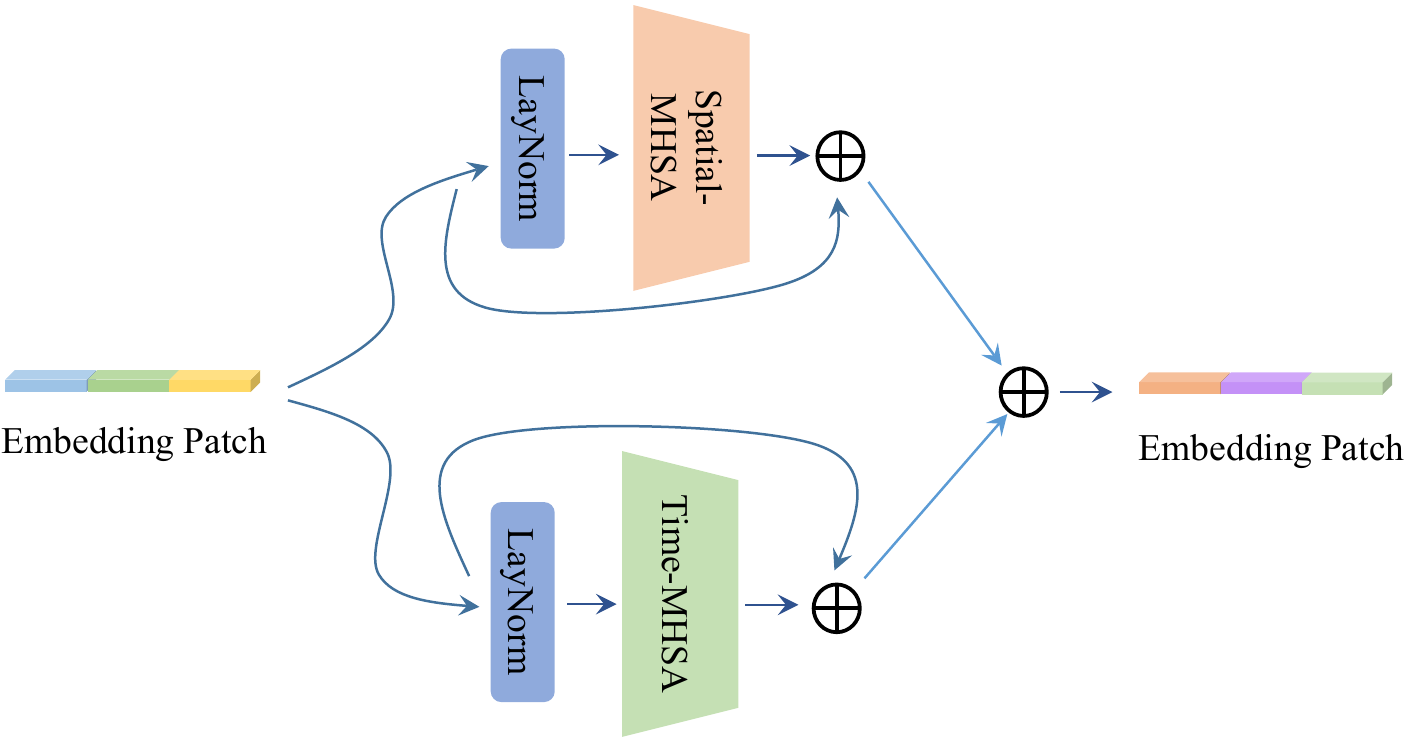}
	\caption{The overall framework of ST-MHSA. ST-MHSA features a parallel dual-branch attention structure for temporal and spatial information, extracting spatial features from relationships across different regions in the frame and temporal features related to environmental changes and head movements of the subjects.}
	\label{figure6}
\end{figure}

ST-MHSA is designed as a multi-faceted attention framework in Global Transformer Branch, concurrently addressing temporal and spatial dimensions. In our analysis, we posit that temporal attributes play a pivotal role in delineating factors like temporal correlation and periodicity crucial. Meanwhile, spatial attributes encompass parameters like facial lighting distribution and facial motion characteristics, which determined the distribution of features extracted from the network. Consequently, we accord equal importance to both temporal and spatial features, advocating for feature extraction methodologies that exhibit parity in efficacy across both domains as shown in Fig. \ref{figure6}.

It is worth noting that in Fig. \ref{figure6}, the multi-head self attention mechanism in terms of time and space is designed as a dual branch structure and provide the abundant spatial-temporal information for the input $\mathbf{X}_{T}^{(k-1)}$ and $\mathbf{X}_{ct}^{(k)}$. $\mathbf{X}_{T}^{(k-1)}$ is the output of the $(k-1)$th Transformer block. $\mathbf{X}_{ct}^{(k)}$ indicates the output of $k$th Conv-Trans Block in CTIM. Before inputting data into Spatial-MHSA and Time-MHSA separately, it is necessary to rearrange $\mathbf{X}_{T}^{(k-1)}$ and $\mathbf{X}_{ct}^{(k-1)}$ into the form fitting Spatial-MHSA and Time-MHSA, which can be expressed as 
\begin{equation}
	\label{eqaution6}
	\begin{split}
		&\mathbf{X}_{TS}^{(k)}=rearrange(\mathbf{X}_{T}^{(k-1)}+\mathbf{X}_{ct}^{(k)})\\
		&\mathbf{X}_{TT}^{(k)}=rearrange(\mathbf{X}_{T}^{(k-1)}+\mathbf{X}_{ct}^{(k)})
	\end{split}
\end{equation}
where $\mathbf{X}_{TS}^{(k)}\in \mathbb{R}^{(B\times nt)\times (nh\times nw)\times D}$ and $\mathbf{X}_{TT}^{(k)}\in \mathbb{R}^{(B\times nh \times nw)\times nt \times D}$ indicate the input of Spatial-MHSA and Time-MHSA respectively. $nt$, $nh$, $nw$ represent the patch numbers obtained by dividing along $T$, $H$, and $W$ of $I$. Furthermore, $nt \times nh \times nw=P$. 

We follow the approach proposed by\cite{arnab2021vivit}, which entails the partitioning of temporal and spatial patches into $\mathbf{X}_{TS}$ and $\mathbf{X}_{TT}$, respectively. This division aims to guide the network towards prioritizing $\mathbf{X}_{TS}$ and $\mathbf{X}_{TT}$ during the implementation of the multi-head self-attention mechanism. It is noteworthy that subsequent to traversing the LayerNorm module and the multi-head self-attention module across temporal and spatial dimensions, we leverage their outputs as spatial-temporal features for the $\mathbf{X}_{T}$. This choice is predicated on the intricate interplay between temporal and spatial features, where processing these intertwined data dimensions in isolation poses the loss of information. Consequently, employing the resultant outputs as attention weights for input modulation serves to deal  with the information loss in decoupling tightly coupled data. In summary, the output of ST-MHSA can be represented as
\begin{equation}
	\label{equation7}
	\begin{split}
	\mathbf{X}_{MT}^{(k)}=&[\mathscr{M}_{S}(\mathscr{L}(\mathbf{X}_{TS}^{(k)}))+\mathbf{X}_{TS}^{(k)}\\
	&+\mathscr{M}_{T}(\mathscr{L}(\mathbf{X}_{TT}^{(k)}))+\mathbf{X}_{TT}^{(k)}]
	\end{split}
\end{equation}
where $\mathscr{M}_{S}$ and $\mathscr{M}_{T}$ indicate the Spatial-MHSA and Time-MHSA respectively and $\mathscr{L}$ denotes the LayerNorm. Subsequently, the output obtained with both temporal and spatial dimensions of information $\mathbf{X}_{MT}^{(k)}$ will be fed into the subsequent steps of the Transformer for further processing.

\subsubsection{Transformer} 
The design of Transformer follows the architecture of \cite{vaswani2017attention}. It is of significance to observe in Fig. \ref{figure7} that the input of the ST-MHSA encompasses not only the output derived from the ST-MHSA module but also incorporates the output generated by the Conv-Trans Block in CTIM. The latter serves the purpose of incorporating pertinent local information into the input of ST-MHSA. Furthermore, it is notable that the architectural design of the FeedForward and LayerNorm follow the principles delineated in the work by \cite{vaswani2017attention}. Therefore, the output of the transformer can be written as
\begin{equation}
	\label{equation8}
	\begin{split}
			&\mathbf{X}_{T}^{(k)}=FFN(\mathscr{L}(\mathbf{X}_{MT}^{(k)}))+\mathscr{L}(\mathbf{X}_{MT}^{(k)})
	\end{split}
\end{equation}
where $FFN$ is the FeedForward block. By passing the Global Transformer Branch and Local Convolution Branch, both global and local features are extracted, thereafter feeding into the rPPG generation module to reconstruct the rPPG signal. Nonetheless, the utilization of distinct branches for global and local feature extraction poses inherent risks of information incompleteness. Moreover, owing to the lack of inductive bias capability of Transformers, Transformer  is susceptible to overfitting on small datasets and convergence challenges. Conversely, convolution operations are born with prior assumptions, but shortage of Global modeling capability in comparison to Transformers. Consequently, by devising information exchange modules between these branches, it is envisaged that the deficiencies in the two branches can be effectively mitigated.

\subsection{CTIM}
\begin{figure}[htbp!]
	\centering
	\includegraphics[scale=0.38]{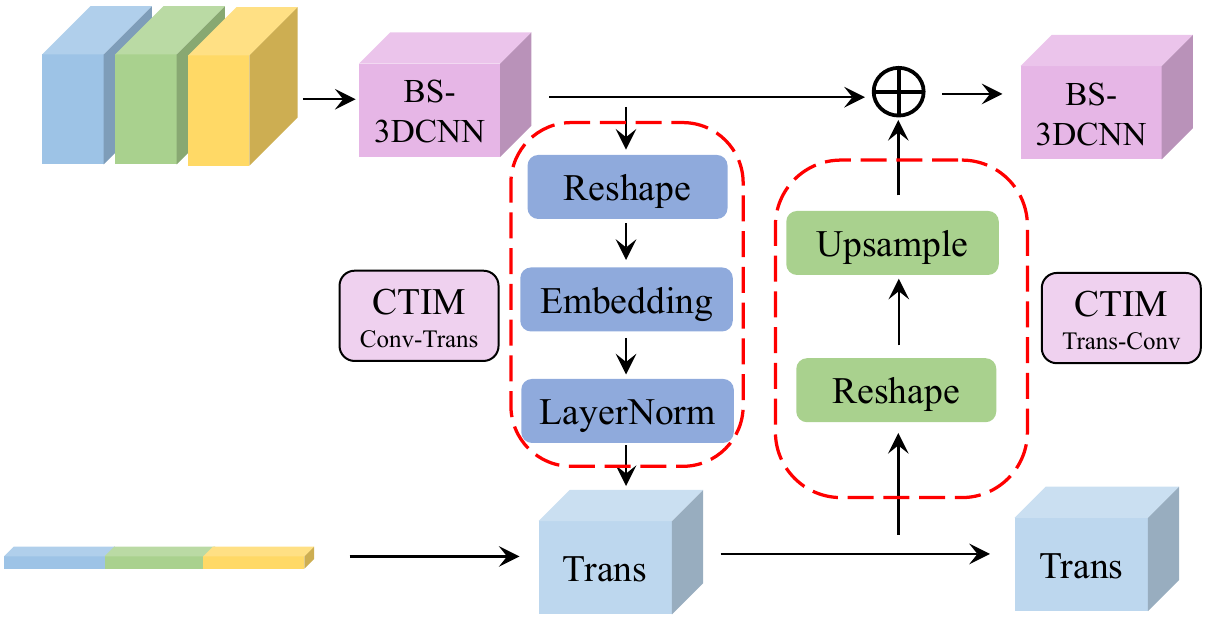}
	\caption{The illustration of CTIM. CTIM enables the interaction and fusion of information extracted by the Local Convolution Branch and the Global Transformer Branch to mitigate the respective limitations of each branch.}
	\label{figure8}
\end{figure}

CTIM is devised to facilitate seamless information exchange between the two branches, aiming to engender a amalgamation of multi-level information and impose inter-branch constraints. It is noteworthy that the inherent challenge associated with extracting global information features through convolutional methodologies is difficlut to solve. This challenge stems from the inherent focus of convolution operations on local correlation information dictated by the convolution kernel size. Conversely, while Transformers excel in global-level modeling, their lack of inherent assumptions akin to those embedded within convolutional paradigms renders their convergence on small datasets intricate. How to effectively overcome the shortcomings of these two is also our original intention in designing CTIM. CTIM includes Trans-Conv Block and Conv-Trans Block as shown in Fig. \ref{figure8}.

\subsubsection{Trans-Conv Block}
The Trans-Conv Block is used to transform the global features captured by the Transformer into a feature form that satisfies the convolution branch, and then weight the features of two branches to add global information to the Local Convolution Branch. 

The dissimilarities in the feature space between the Global Transformer  Branch and Local Convolution Branch necessitate an dimension transformation of their respective feature dimension. Notably, during the data initialization phase, the dimensions $T$, $H$, and $W$ of the input data are rendered into distinct cubes due to the Patch operation and Embedding processes. Consequently, reshaping the Embedding features into tensors with dimensions corresponding to $T$, $H$, and $W$ becomes necessary. Nonetheless, it is noteworthy that the channel dimensional of the data remains unaltered throughout this transformation process. As a consequence, the number of channels pertaining to the features within the transformed Global Transformer Branch should align with those within the Local Convolution Branch namely ${}^{t}\mathbf{X}_{TC}^{(k)}=Reshape(\mathbf{X}_{T}^{(k-1)})$, ${}^{t}\mathbf{X}_{TC}^{(k)}\in \mathbb{R}^{B\times C\times T\times H \times W}$ and $\mathbf{X}_{T}^{(k-1)}\in \mathbb{R}^{B\times P \times D}$. 

Subsequent to the reshaping process, it becomes apparent that the dimensions $T$, 
$H$, and $W$ of the resultant features are considerably smaller in comparison to those of the Local Convolution Branch features. Consequently, it becomes imperative to engage in an upsampling operation to augment the dimensions of the reshaped features. Through this upsampling operation,the sizeof the features derived from the Global Transformer Branch are equivalent those of the features from the Local Convolution Branch.  Nevertheless, it is pertinent to note that upsampling, being an interpolation technique, may bring the information redundancy. In light of this consideration, following the upsampling phase, a 3D convolution utilizing a kernel size of 1 is employed in BS-3DCNN block for further refinement. Subsequently, these features from CTIM are added with those of the Local Convolution Branch, culminating in their fusion for subsequent processing. The output of the final CTIM Trans Conv Block can be represented as
\begin{equation}
	\label{equation9}
		\mathbf{X}_{tc}^{(k)}=Upsample(Reshape(\mathbf{X}_{T}^{(k)}))
\end{equation}
where $Upsample$ indicates the upsample operation. Subsequently, the output of the Trans Conv Block in CTIM will be sent to the Local Convolution Branch.

\subsubsection{Conv-Trans Block}
The Conv-Trans Block serves the purpose of harmonizing the feature dimensions originating from the Local Convolution Branch with those characteristic of the Global Transformer Branch. Additionally, it facilitates the integration of local information gained from convolution with the global features of the Transformer. Such integration helps the Transformer posing with an inductive bias, thereby achieving effective convergence. The Local Revolution Branch boasts a larger feature size than the Global Transformer Branch, endowing it with richer local information. Nevertheless, this abundance of local detail from the Local Convolution Branch might be redundancy in the Global Transformer Branch. To address this issue, we undertook a feature reshape within the Conv-Trans Block. Utilizing linear layers for embedding, we readjust the features to better align with the information distribution needs of the Global Transformer Branch. In addition, layer normalization is used to standardize the output of Conv Trans Blocks. After the above design, the output of Conv Trans Block can be written as
\begin{equation}
	\label{equation10}
	\mathbf{X}_{ct}^{(k)}=\mathscr{L}(Lin(Reshape(\mathbf{X}_{C}^{(k-1)}))).
\end{equation}
The output $\mathbf{X}_{ct}^{(k)}$ of the final Conv-Trans Block will be fed into the Transformer Block. After multi-layer feature extraction, the extracted features will be sent to RGM for the final generation of BVP. 

\subsection{RGM}
\begin{figure}[htbp!]
	\centering
	\includegraphics[scale=0.37]{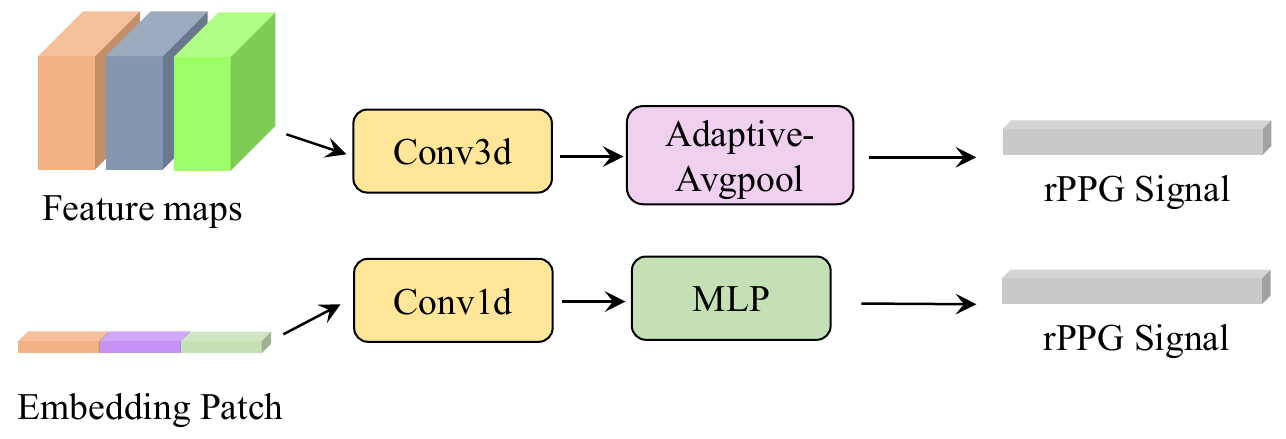}
	\caption{The overall framework of RGM.}
	\label{figure9}
\end{figure}

RGM is used as a downstream task module for the generation of BVP and the structure can be seen in the Fig. \ref{figure9}. 
RGM generates BVP based on the features extracted from Local Convolution Branch and Global Transformer Branch, respectively. Given the invariant shape of the T-dimension across the features within the Local Convolution Branch, we employ a 3-dimensional convolution to reduce the number of feature channels while simultaneously amalgamating the spatial dimensions via 2-dimensional adaptive average pooling, ultimately obtaining the rPPG signal. For the Global Transformer Branch, we diminish the feature channel of the output from the Global Transformer Branch by leveraging 1-Dimension convolution. This is followed by feature flattening and rPPG signal facilitated by Multilayer Perceptron (MLP). Ultimately, after undergoing RGM, we have
\begin{equation}
	\label{equation11}
	\begin{split}
		{R}_{1}=&\psi(\varsigma_{1}(\mathbf{X}^{(k)}_{C}))\\
		{R}_{2}=&Lin(\varsigma_{2}(\mathbf{X}^{(k)}_{T}))
	\end{split}
\end{equation}
where $\psi$ is the adaptive Avg-pool function, $R_1$ and $R_2$ are the estimated BVP signals of RGM with the features originating from Local Convolution Branch and Global Transformer Branch. Subsequently, the obtained $R_{1}$ and $R_2$ will be used for network optimization, HR measurement and HRV estimation.
 
\subsection{Network Optimization}
To recover the rPPG signal with accurate systematic peak instants, the negative Pearson coefficient is used as a partial loss $L_{p}$ in network optimization to gradually approximate the actual rPPG signal, following \cite{yu2022physformer,niu2019rhythmnet}. In addition, the combination of negative Pearson coefficient loss $L_p$ and Smooth L1 loss $L_1$ is used as the overall loss for network optimization\cite{yu2019remote,song2021pulsegan}, which can be represented as
\begin{equation}
	\label{equation12}
	L_p=1-\frac{\sum_{i=1}^{l}(y_i-\bar{y})(Y_i-\bar{Y})}{\sqrt{\sum_{i=1}^{l}(y_i-\bar{y})^2}\sqrt{\sum_{i=1}^{l}(Y_i-\bar{Y})^2}}
\end{equation}
where $y_i$ and $Y_i$ present the $i$th sampling points of the predict rPPG signal and the true rPPG siganl. And $l$ is the length of rPPG signal. In addition, $\bar{y}$ and $\bar{Y}$ indicate the mean value of the predict rPPG signal and the true rPPG siganl. And the $L_1$ can be expressed as
\begin{equation}
	\label{equation13}
	L_1=\left\{
	\begin{aligned}
		0.5(Y_i-y_i)^2 &,   & (Y_i-y_i)<1  \\
		|(Y_i-y_i)|-0.5&,   & otherwise
	\end{aligned}
	\right.
\end{equation}
Eventually, $L_p$ and $L_1$ are combined to the final loss $L$ 
\begin{equation}
	\label{equation14}
	L=\alpha*L_p+(1-\alpha)*L_1
\end{equation}
where $\alpha$ is a hyper-parameter to balance the $L_p$ and $L_1$ and $\alpha=0.5$ in this paper. It is worth noting that VidFormer will simultaneously output estimated BVP signals from both Local Convolution Branch and Global Transformer Branch. Therefore, it is necessary to optimize $R_1$ and $R_2$ separately, and average the output HR from $R_1$ and $R_2$ as the final HR.

\section{Experiments}
\subsection{Datasets}
 We conduct our experiments on five public datasets: UBFC-rPPG\cite{bobbia2019unsupervised}, PURE\cite{stricker2014non}, DEAP\cite{koelstra2011deap}, ECG-fitness\cite{vspetlik2018visual}, and COHFACE\cite{heusch2017reproducible}. 
 \subsubsection{UBFC-rPPG}The UBFC-rPPG dataset contains 42 videos and corresponding recorded PPG signals. This dataset collected the heartbeat and videos of 42 health subjects playing a time-sensitive mathematical game under varying amounts of ambient light. The videos were recorded at 30 fps with a resolution of 640 $\times$ 480 in the uncompressed 8-bit RGB format using a low-cost Logitech C920 HDpro web camera. The ground truth PPG signals were acquired using a CMS50E transmission pulse oximeter at a sampling rate of 30 Hz.
 \subsubsection{PURE}The PURE dataset recorded 60 facial videos of 10 health subjects. These subjects were asked to perform six types of head movements in front of the camera for one minute. These videos were collected using a camera with a resolution of 640 $\times$ 480 and a frame rate of 30 fps, and the ground truth PPG signal was collected using a pulse oximeter with a sampling rate of 60 Hz. We down-sampled the BVP signal of the PURE dataset to 30 Hz.
 \subsubsection{DEAP} The DEAP dataset consists of 874 facial videos associated with multi channel physiological signals. They were taken from 22 subjects by playing one-minute musical excerpts to them. Each facial video has a resolution of 720 × 576 and a frame rate of 50 fps. And the ground truth PPG signals is obtained by sampling at a 512 Hz sampling rate. Therefore, we down-sampled the PPG signals to 30 Hz before our module training.
 \subsubsection{ECG-fitness} The ECG-fitness dataset was recorded from 17 subjects while performing different physical activities. The facial videos were recorded by Logitech camera with the resolution of 1920 $\times$ 1080 and a frame rate of 30 fps. ECG signals were recorded simultaneously with the videos using a two lead Viatom CheckMeTMPro device with CC5 lead owning a sampling rate of 125 Hz. We followed the work of \cite{das2023time} to generate the synthetic PPG signals with sample rate of 30 Hz.
 \subsubsection{COHFACE}The COHFACE dataset records facial videos and PPG signals of 40 health subjects under different natural lighting conditions. The videos were recorded at 20 fps with a resolution of 640 $\times$ 480 using a Logitech HDC525 camera. Simultaneously, the PPG signals were acquired at 256 Hz. Moreover, we down-sampled the PPG signals to 40 Hz.
 
\subsection{Implementation Details}
For facial videos from five datasets, we randomly select one segment and slice it using a window with length of 250 frames and a step size of 50 frames. Meanwhile, we scale the sliced video data to a resolution of $128\times128$ and optimize it by sending it to the network with batch size equal to 2. For the Global Transformer Branch, training video data will be divided into patches of size $25 \times16\times16$. The extraction of Region of Interes (ROI) of Facial video processing is achieved by the utilization of the dlib library for the detection of facial keypoints and subsequent computation of the facial center point for the region cropping. 

Our model is trained for 150 epochs using one single NVIDIA GeForce RTX 4090 and PyTorch 1.13\cite{paszke2019pytorch} on UBFC-rPPG, PURE and 100 epochs on COHFACE, ECG-fitness. For DEAP, our model is trained for 30 epochs. We use the AdamW optimizer\cite{loshchilov2017decoupled} and Cosine Annealing Warm Restart learning rate adjustment strategy to adjust the learning rate. The maximum learning rate and the Initial learning rate are set to $8\times 10^{-5}$, the minimum is set to $2\times 10^{-9}$, and the batch size is set to 2. Meanwhile, the weight decay is set to $5\times 10^{-4}$.

\subsection{Evaluation Protocol}
\label{section-c}
\cite{park2022self,botina2022rtrppg,chen2018deepphys} utilized estimated BVP signal to calculate HR, RF, and HRV. These estimated values were then compared with corresponding ground truth measurements to evaluate the performance of the network. We follow their methods to calculate HR and HRV using estimated BVP on five datasets UBFC, PURE, COHFACE, ECG fitness, and DEAP. Moreover, we conducted cross dataset testing between datasets UBFC, PURE and COHFACE to validate the effectiveness of our proposed model. The calculation method for HR, RF and HRV is implemented using the HeartPy toolbox\cite{van2019analysing}.

We follow the work of [15], [20] to use mean absolute error (MAE), root mean square error (RMSE) and Pearson’s correlation coefficient ($r$) as evaluation metrics for HR. For HRV, we follow \cite{poh2010advancements,li2018obf} calculating the three attributes of HRV, i.e., low frequency (LF), high frequency (HF), and the LF/HF ratio. The results of LF and HF are obtained by calculating the interbeat intervals of BVP under the low frequency (0.04 Hz to 0.15 Hz) and high frequency (0.15 Hz to 0.4 Hz) bands. For each attributes of HRV, we follow \cite{niu2020video,lu2021dual,yu2019remote} to employ standard deviation (STD), RMSE and $r$ as evaluation metrics. For RF, we also report the STD, RMSE and $r$ as per most comparable methods\cite{verkruysse2008remote,wang2016algorithmic,niu2020video,yu2019remote,yu2022physformer,lu2021dual,yue2023facial,gideon2021way,sun2022contrast,de2013robust}. In the context of MAE, RMSE, and STD, smaller values indicate better estimated HR result, whereas for $r$, the closer the value of $r$ is to $1$, the better the result.

\subsection{Results}
\subsubsection{HR Evaluation}
\begin{table*}[t]
	\centering
	\caption{Intra-dataset: Comparison to the State of the Art on HR Estimation}
	\label{table1}
	\begin{tabular}{p{26pt}p{59pt}p{20pt}p{22pt}m{9pt}p{20pt}p{22pt}p{9pt}p{20pt}p{22pt}p{9pt}p{20pt}p{22pt}p{9pt}p{20pt}p{22pt}c}		
		\toprule
		\multirow{2}{*}{\makecell[c]{Method-\\type}} &\multirow{2}{*}{Method}  &\multicolumn{3}{c}{UBFC-rPPG}  &\multicolumn{3}{c}{PURE} &\multicolumn{3}{c}{COHFACE} &\multicolumn{3}{c}{ECG-fitness} &\multicolumn{3}{c}{DEAP}\\
		\cmidrule(r){3-5} \cmidrule(r){6-8} \cmidrule(r){9-11} \cmidrule(r){12-14} \cmidrule(r){15-17}
		~ &~ &\makecell[c]{MAE\\(bpm)} &\makecell[c]{RMSE\\(bpm)} &$r$ &\makecell[c]{MAE\\(bpm)} &\makecell[c]{RMSE\\(bpm)} &$r$ &\makecell[c]{MAE\\(bpm)} &\makecell[c]{RMSE\\(bpm)} &$r$ &\makecell[c]{MAE\\(bpm)} &\makecell[c]{RMSE\\(bpm)} &$r$ &\makecell[c]{MAE\\(bpm)} &\makecell[c]{RMSE\\(bpm)} &$r$\\
		\midrule
		\multirow{7}{*}{\makecell[c]{Tradi-\\tional}} &GREEN\cite{verkruysse2008remote} &$6.01$ &$7.87$ &$0.29$ &$4.39$ &$11.60$ &$0.90$ &$-$ &$-$ &$-$ &$-$ &$-$ &$-$ &$8.10$ &$11.17$ &$0.80$\\
		~ &ICA\cite{poh2010advancements} &$5.42$ &$10.47$ &$0.82$ &$5.66$ &$11.15$ &$0.89$ &$7.61$ &$9.65$ &$0.72$ &$22.73$ &$26.82$ &$0.28$ &$-$ &$-$ &$-$\\
		~  &POS\cite{wang2016algorithmic} &$4.05$ &$8.75$ &$0.78$ &$3.14$ &$10.57$ &$0.95$ &$6.58$ &$11.90$ &$0.49$ &$23.46$ &$30.60$ &$0.12$ &$7.39$ &$10.25$ &$0.82$\\ 
		~ &CHROM\cite{de2013robust} &$8.20$ &$9.92$ &$0.27$ &$3.82$ &$6.80$ &$0.97$ &$9.70$ &$10.80$ &$0.67$ &$19.04$ &$23.84$ &$0.10$ &$7.47$ &$10.31$ &$0.82$\\
		~ &2SR\cite{wang2015novel} &$6.90$ &$18.50$ &$0.65$ &$2.44$ &$3.06$ &$0.98$ &$20.98$ &$25.84$ &$0.32$ &$43.66$ &$-$ &$-$ &$-$ &$-$ &$-$\\
		~ &PBV\cite{de2014improved} &$5.82$ &$6.68$ &$0.82$ &$-$ &$-$ &$-$ &$8.20$ &$10.85$ &$0.03$ &$24.53$ &$28.35$ &$0.22$ &$-$ &$-$ &$-$\\
		~ &LiCVPR\cite{li2014remote} &$-$ &$-$ &$-$ &$28.22$ &$-$ &$0.38$ &$19.98$ &$-$ &$0.44$ &$63.25$ &$-$ &$0.02$ &$-$ &$-$ &$-$\\
		\midrule
		\multirow{15}{*}{\makecell[c]{Super-\\vised}}
		~ &HRCNN\cite{vspetlik2018visual}$\blacktriangle$ &$-$ &$-$ &$-$ &$1.84$ &$2.37$ &$0.98$ &$8.10$ &$-$ &$0.29$ &$14,48$ &$-$ &$0.50$ &$-$ &$-$ &$-$\\
		~ &SynRhythm\cite{niu2018synrhythm}$\star$ &$5.59$ &$6.82$ &$0.72$ &$2.71$ &$4.86$ &$0.98$ &$-$ &$-$ &$-$ &$-$ &$-$ &$-$ &$5.08$ &$5.92$ &$0.87$\\
		~ &Meta-rppg\cite{lee2020meta}$\blacktriangle$ &$5.97$ &$7.42$ &$0.53$ &$2.52$ &$4.63$ &$0.98$ &$-$ &$-$ &$-$ &$-$ &$-$ &$-$ &$5.16$ &$6.00$ &$0.87$\\
		~ &DeepPhys\cite{chen2018deepphys}$\blacktriangle$ &$4.58$ &$14.76$ &$0.78$ &$-$ &$-$ &$-$ &$6.89$ &$13.89$ &$0.34$ &$-$ &$-$ &$-$ &$5.90$ &$7.43$ &$0.87$\\
		~ &PhysNet\cite{yu2019remote}$\blacktriangle$ &$2.95$ &$3.67$ &$0.98$ &$1.90$ &$3.44$ &$0.98$ &$5.38$ &$10.76$ &$0.87$ &$17.33$ &$18.62$ &$0.43$ &$-$ &$-$ &$-$\\
		~ &Rhythmnet\cite{niu2019rhythmnet}$\star$ &$1.19$ &$2.10$ &$0.98$ &$-$ &$-$ &$-$ &$-$ &$-$ &$-$ &$-$ &$-$ &$-$ &$7.47$ &$8.96$ &$0.82$\\
		~ &rPPGNet\cite{yu2019remote}$\blacktriangle$ &$-$ &$-$ &$-$ &$-$ &$-$ &$-$ &$-$ &$-$ &$-$ &$-$ &$-$ &$-$ &$6.22$ &$7.73$ &$0.83$\\
		~ &AND-rPPG\cite{lokendra2022and}$\star$ &$2.67$ &$4.07$ &$0.96$ &$-$ &$-$ &$-$ &$3.82$ &$5.10$ &$0.79$ &$-$ &$-$ &$-$ &$-$ &$-$ &$-$\\
		~ &Das\cite{das2023time}$\blacktriangle$ &$0.57$ &$0.77$ &$\mathbf{0.99}$ &$-$ &$-$ &$-$ &$0.99$ &$1.98$ &$0.96$ &$4.27$ &$5.44$ &$0.94$ &$-$ &$-$ &$-$\\
		~ &PulseGAN\cite{song2021pulsegan}$\star$ &$1.19$ &$2.10$ &$0.98$ &$2.28$ &$4.29$ &$\mathbf{0.99}$ &$-$ &$-$ &$-$ &$-$ &$-$ &$-$ &$4.86$ &$5.70$ &$0.88$\\
		~ &CPulse\cite{mehta2023cpulse}$\star$ &$1.06$ &$1.48$ &$\mathbf{0.99}$ &$0.98$ &$1.94$ &$\mathbf{0.99}$ &$1.99$ &$3.83$ &$0.94$ &$-$ &$-$ &$-$ &$-$ &$-$ &$-$\\
		~ &PFE-TFA\cite{li2023learning}$\blacktriangle$ &$0.76$ &$1.62$ &$\mathbf{0.99}$ &$1.44$ &$2.50$ &$0.98$ &$1.31$ &$3.92$ &$0.96$ &$-$ &$-$ &$-$ &$-$ &$-$ &$-$\\
		~ &Dual-GAN\cite{lu2021dual}$\star$ &$0.44$ &$\mathbf{0.67}$ &$\mathbf{0.99}$ &$0.82$ &$1.31$ &$\mathbf{0.99}$ &$-$ &$-$ &$-$ &$-$ &$-$ &$-$ &$3.25$ &$4.11$ &$0.91$\\
		~ &Physformer\cite{yu2022physformer}$\blacktriangle$ &$0.40$ &$0.71$ &$\mathbf{0.99}$ &$1.10$ &$1.75$ &$\mathbf{0.99}$ &$-$ &$-$ &$-$ &$-$ &$-$ &$-$ &$3.03$ &$3.96$ &$0.92$\\
		~ &\textbf{VidFormer(ours)}$\blacktriangle$ &$\mathbf{0.32}$ &$0.97$ &$\mathbf{0.99}$ &$\mathbf{0.42}$ &$\mathbf{1.18}$ &$\mathbf{0.99}$ &$\mathbf{0.53}$ &$\mathbf{1.31}$ &$\mathbf{0.99}$ &$\mathbf{0.64}$ &$\mathbf{2.16}$ &$\mathbf{0.99}$ &$\mathbf{0.75}$ &$\mathbf{1.46}$ &$\mathbf{0.98}$\\
		\midrule
		\multirow{6}{*}{\makecell[c]{Unsup-\\ervised}}
		~ &Gideon\cite{gideon2021way}$\blacktriangle$ &$1.85$ &$4.28$ &$0.93$ &$2.23$ &$2.97$ &$0.99$ &$3.6$ &$4.6$ &$\colorbox{lightgray}{0.95}$ &$-$ &$-$ &$-$ &$5.13$ &$6.16$ &$0.86$\\
		~ &SiNC\cite{speth2023non}$\blacktriangle$ &$0.59$ &$1.83$ &$\colorbox{lightgray}{0.99}$ &$\colorbox{lightgray}{0.61}$ &$1.84$ &$\colorbox{lightgray}{1.00}$ &$-$ &$-$ &$-$ &$-$ &$-$ &$-$ &$-$ &$-$ &$-$\\
		~ &Sun\cite{sun2022contrast}$\blacktriangle$ &$0.64$ &$1.00$ &$\colorbox{lightgray}{0.99}$ &$1.00$ &$\colorbox{lightgray}{1.40}$ &$0.99$ &$4.93$ &$7.71$ &$0.73$ &$\colorbox{lightgray}{3.27}$ &$\colorbox{lightgray}{4.86}$ &$\colorbox{lightgray}{0.92}$ &$-$ &$-$ &$-$\\
		~ &SLF-RPM\cite{yang2022simper}$\blacktriangle$ &$8.39$ &$9.70$ &$0.70$ &$-$ &$-$ &$-$ &$-$ &$-$ &$-$ &$-$ &$-$ &$-$ &$-$ &$-$ &$-$\\
		~ &Zhang\cite{zhang2024self}$\blacktriangle$ &$1.88$ &$2.94$ &$0.98$ &$1.70$ &$2.38$ &$0.99$ &$\colorbox{lightgray}{1.49}$ &$\colorbox{lightgray}{4.55}$ &$0.90$ &$-$ &$-$ &$-$ &$-$ &$-$ &$-$\\
		~ &Yue\cite{yue2023facial}$\blacktriangle$ &$\colorbox{lightgray}{0.58}$ &$\colorbox{lightgray}{0.94}$ &$\colorbox{lightgray}{0.99}$ &$1.23$ &$2.01$ &$0.99$ &$-$ &$-$ &$-$ &$-$ &$-$ &$-$ &$\colorbox{lightgray}{4.20}$ &$\colorbox{lightgray}{5.18}$ &$\colorbox{lightgray}{0.90}$\\
		\bottomrule
	\end{tabular}
\begin{tablenotes}
	\item $\star$ and $\blacktriangle$ denote the non-end-to-end and end-to-end DeepLearning-based Method respectively. The best supervision approach is marked in bold. The best Unsupervised approach is marked in shadow.
\end{tablenotes}
\end{table*}
First, we conducted intra-dataset testing across five datasets, comparing our evaluation results with those of various methodologies. These methodologies encompass a spectrum of approaches, ranging from conventional to supervised and unsupervised techniques as detailed in Table \ref{table1}. 

Notably, our comparative analysis underscores a pronounced discrepancy between traditional methodologies and those rooted in deep learning paradigms. This disparity arises from the inherent limitations of traditional methods, which often hinge on predefined assumptions that may inadequately capture the intricacies of diverse environments. Furthermore, the efficacy of traditional feature extraction techniques is impeded by constraints pertaining to information density of features and the presumptions underlying feature selection, thereby impeding the reconstruction of BVP signals. The DNN-based approach harnesses extensive datasets to optimize the model effectively. Through the establishment of a sophisticated solution space and the iterative refinement facilitated by gradient descent methods, these techniques enable the extraction of intricate and high-dimensional features. Consequently, DNN-based methods demonstrate efficacy in discerning BVP signals across varied environmental contexts, thereby manifesting superior generalization capabilities.

\begin{figure}[htbp!]
	\centering
	\includegraphics[scale=0.3]{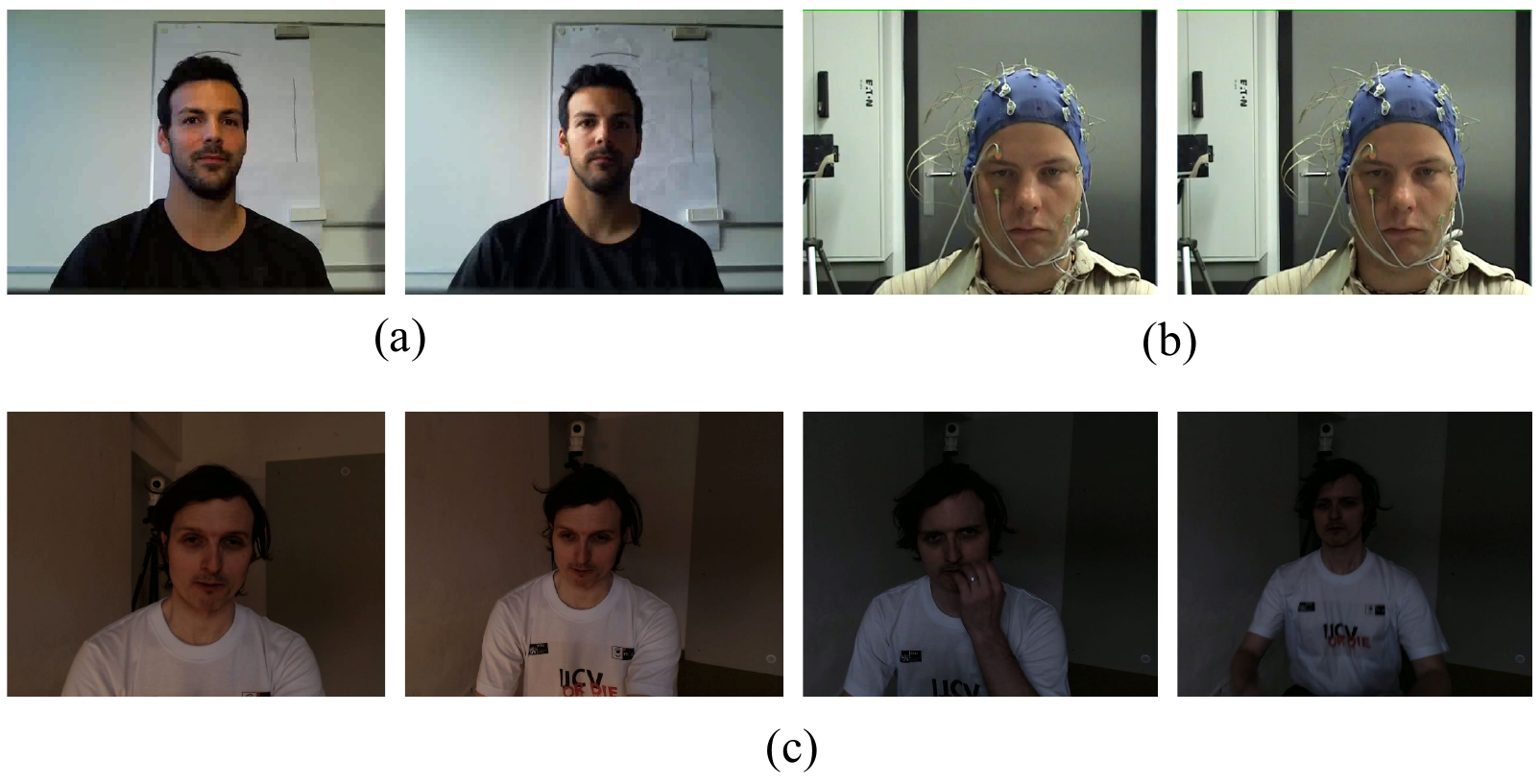}
	\caption{Face images in COHFACE, DEAP and ECG-fitness, (a) denotes the images in COHFACE, (b) indicate the images in DEAP and (c) are the images in ECG-fitness.}
	\label{figure10}
\end{figure}

Moreover, the unsupervised methods delineated in the table have achieved a level of performance commensurate with their supervised counterparts. However, it is worht noting that the MAE of SOTA models in UBFC and PURE has dwindled to decrease 1 bpm, yet a conspicuous gap persists across the COHFACE, ECG-fitness, and DEAP datasets. This discrepancy arises primarily from the disparate sizes of the datasets, with UBFC-rPPG and PURE featuring considerably smaller data volumes compared to COHFACE, ECG-fitness, and DEAP. Consequently, the latter datasets entail more intricate lighting conditions and diverse test subjects. Specifically, the COHFACE dataset encompasses multiple facial video sequences captured under varying natural lighting conditions, as visually depicted in Fig. \ref{figure10} (a). The ECG-fitness dataset incorporates ECG signals from participants in diverse motion states and lighting conditions, so the extraction process is significantly impeded by motion artifacts as shown in Fig. \ref{figure10} (b). Likewise, the DEAP dataset comprises $26055$ face videoes. The subjects is necessitating the wearing of electrode patches on facial regions for human signal feature extraction. Therefore, this renders the employment of cameras for BVP signal acquisition challenging, as depicted in the Fig. \ref{figure10} (c).

Our proposed method exhibits superior performance compared to traditional, supervised and unsupervised approaches. Notably, it achieves SOTA levels of accuracy on the PURE, COHFACE, ECG-fitness, and DEAP datasets. On the UBFC-rPPG dataset, our method achieves a remarkable MAE metric of 0.32 bpm, representing a reduction of 0.08 bpm compared to the Physfomer baseline. Moreover, RMSE metric on UBFC-rPPG surpasses the majority of both supervised and unsupervised approaches. It is particularly noteworthy that VidFormer has the capability to achieve MAE levels below 1 bpm on large datasets such as DEAP and datasets with complex recording environments like ECG-fitness. This capability is particularly significant as it opens avenues for non-contact detection of human vital signs in diverse and challenging environments. Additionally, it is observed that VidFormer experiences a slight increase in MAE with the gradual expansion of dataset size. However, this marginal increase can be ignored when the growth of dataset size increase.

\begin{figure*}[t]
	\centering
	\includegraphics[scale=0.50]{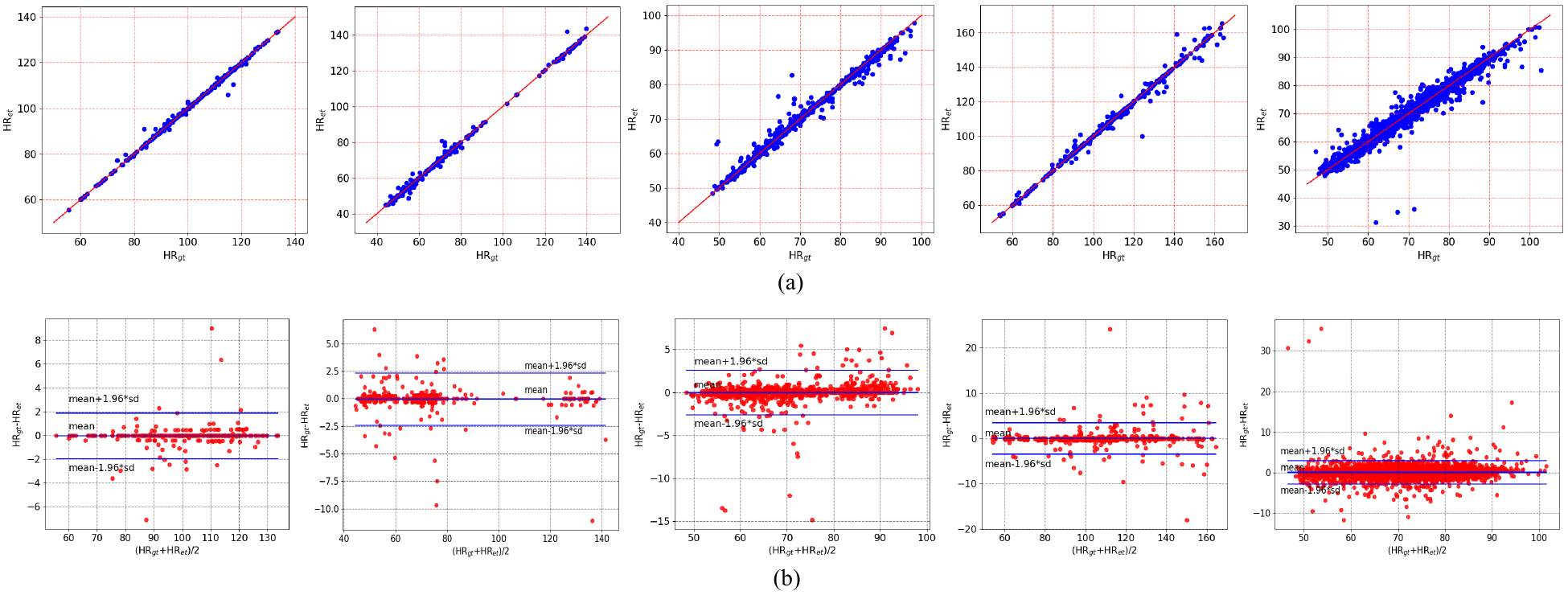}
	\caption{(a) is the scatter plots between ground-truth HR and estimated HR on UBFC-rPPG, PURE, COHFACE, ECG-fitness and DEAP. The straight red line is a fitting line computed by all points. HR$_{et}$ and HR$_{gt}$ indicate the estimate HR and the ground truth HR, respectively. (b) is the Bland-Altman plots of the HR estimation error on UBFC-rPPG, PURE, COHFACE, ECG-fitness and DEAP. The blue lines represent the mean and $95\%$ limits of agreement, respectively.}
	\label{figure11}
\end{figure*}

In addition, we conducted Bland-Altman plot and scatter plots on the results of five datasets as depicted in Fig. \ref{figure11}.
The analysis of Fig. \ref{figure11} reveals a well association between the HR$_{et}$ and HR$_{gt}$, indicating a robust correlation. VidFormer demonstrates reliable effectiveness in predicting HR across a spectrum spanning from 44 bpm to 160 bpm. It is noteworthy that the high HR recorded in the ECG-fitness dataset stem from the subjects undergoing physical exercise, thereby yielding comparatively higher HR among the other four datasets. Furthermore, VidFormer exhibits substantially elevated Pearson correlation coefficients between HR$_{et}$ and HR$_{gt}$ across the COHFACE, ECG fitness, and DEAP datasets compared to alternative methodologies.

\subsubsection{RF and HRV Evaluation}
We further conduct experiments for RF and HRV estimation on the UBFC-rPPG dataset and HRV is represented by its three attributes (LF, HF, LF/HF) as mentioned in subsection \ref{section-c}. Similar to the HR evaluation, we compare our approach with SOTA methods \cite{wang2016algorithmic}, \cite{de2013robust}, \cite{verkruysse2008remote}, \cite{niu2020video}, \cite{yu2019remote}, \cite{song2021pulsegan}, \cite{gideon2021way}, \cite{sun2022contrast}, \cite{yue2023facial} as indicate in Table \ref{table2}.
\begin{table*}[t]
	\centering
	\caption{Comparison to State of the Art on RF and HRV Estimation}
	\label{table2}
	\begin{tabular}{cccccccccccccc}
		\toprule
		\multirow{2}{*}{\makecell[c]{Method-type}} &\multirow{2}{*}{\makecell[c]{Method}} &\multicolumn{3}{c}{LF(n.u.)} &\multicolumn{3}{c}{HF(n.u.)} &\multicolumn{3}{c}{LF/HF} &\multicolumn{3}{c}{RF(Hz)}\\
		\cmidrule(r){3-5} \cmidrule(r){6-8} \cmidrule(r){9-11} \cmidrule(r){12-14}
		~ & ~& STD& RMSE& $r$& STD& RMSE& $r$& STD& RMSE& $r$& STD& RMSE& $r$\\
		\midrule
		\multirow{3}{*}{Traditional} & CHROM\cite{de2013robust} &$0.243$ &$0.240$ &$0.159$ &$0.243$ &$0.240$ &$0.159$ &$0.655$ &$0.645$ &$0.226$ &$0.086$ &$0.089$ &$0.102$\\
		~ & GREEN\cite{verkruysse2008remote} &$0.186$ &$0.186$ &$0.280$ &$0.186$ &$0.186$ &$0.280$ &$0.361$ &$0.365$ &$0.492$ &$0.087$ &$0.086$ &$0.111$\\
		~ & POS\cite{wang2016algorithmic} &$0.171$ &$0.169$ &$0.479$ &$0.171$ &$0.169$ &$0.479$ &$0.405$ &$0.399$ &$0.518$ &$0.109$ &$0.107$ &$0.087$\\
		\midrule
		\multirow{5}{*}{Supervised} &CVD \cite{niu2020video}$\star$ &$0.053$ &$0.065$ &$0.740$ &$0.053$ &$0.065$ &$0.740$ &$0.169$ &$0.168$ &$0.812$ &$0.017$ &$0.018$ &$0.252$\\
		~ &rPPGNet\cite{yu2019remote}$\blacktriangle$ &$0.071$ &$0.070$ &$0.686$  &$0.071$ &$0.070$ &$0.686$ &$0.212$ &$0.208$ &$0.744$ &$0.030$ &$0.034$ &$0.233$\\
		~ &Dual-GAN\cite{song2021pulsegan}$\star$ &$0.034$ &$0.035$ &$0.891$ &$0.034$ &$0.035$ &$0.891$ &$0.131$ &$0.136$ &$0.881$ &$0.010$ &$0.010$ &$0.395$\\
		~ &Physformer\cite{yu2022physformer}$\blacktriangle$ &$0.030$ &$0.032$ &$0.895$  &$0.030$ &$0.032$ &$0.895$ &$0.126$ &$0.130$ &$0.893$ &$0.009$ &$\mathbf{0.009}$ &$0.413$\\
		~ &\textbf{VidFormer(ours)}$\blacktriangle$ &$\mathbf{0.027}$ &$\mathbf{0.028}$ &$\mathbf{0.929}$ &$\mathbf{0.027}$ &$\mathbf{0.028}$ &$\mathbf{0.929}$ &$\mathbf{0.095}$ &$\mathbf{0.101}$ &$\mathbf{0.928}$ &$\mathbf{0.008}$ &$\mathbf{0.009}$ &$\mathbf{0.415}$\\
		\midrule
		\multirow{3}{*}{Unsupervised} &Gideon\cite{gideon2021way}$\blacktriangle$ &$0.091$ &$0.139$ &$0.694$  &$0.091$ &$0.139$ &$0.694$ &$0.525$ &$0.691$ &$0.684$ &$0.061$ &$0.098$ &$0.103$\\
		~ & Contras-Phys\cite{sun2022contrast}$\blacktriangle$ &$0.050$ &$0.098$ &$0.798$ &$0.050$ &$0.098$ &$0.798$ &$0.205$ &$0.395$ &$0.782$ &$0.055$ &$0.083$ &$0.347$\\
		~ & Yue\cite{yue2023facial}$\blacktriangle$ &$0.047$ &$0.062$ &$0.769$  &$0.047$ &$0.062$ &$0.769$ &$0.160$ &$0.164$ &$0.831$ &$0.023$ &$0.028$ &$0.351$\\
		\bottomrule
	\end{tabular}
\begin{tablenotes}
	\item $\star$ and $\blacktriangle$ denote the non-end-to-end and end-to-end DeepLearning-based Method respectively. The best supervision approach is marked in bold. The best Unsupervised approach is marked in shadow.
\end{tablenotes}
\end{table*}

It is pertinent to highlight that our method produces very competitive results on par with SOTA methods in HRV estimation. This achievement underscores the effectiveness of VidFormer in accurately reconstructing rPPG signals, particularly in determining systolic peak values with precision. Such capability positions VidFormer as a viable solution for diverse applications, including but not limited to facial expression recognition\cite{wu2023recognizing,li2020deep} and healthcare domains\cite{shi2019atrial,yan2018contact}.

\begin{figure*}[t]
	\centering
	\includegraphics[scale=0.65]{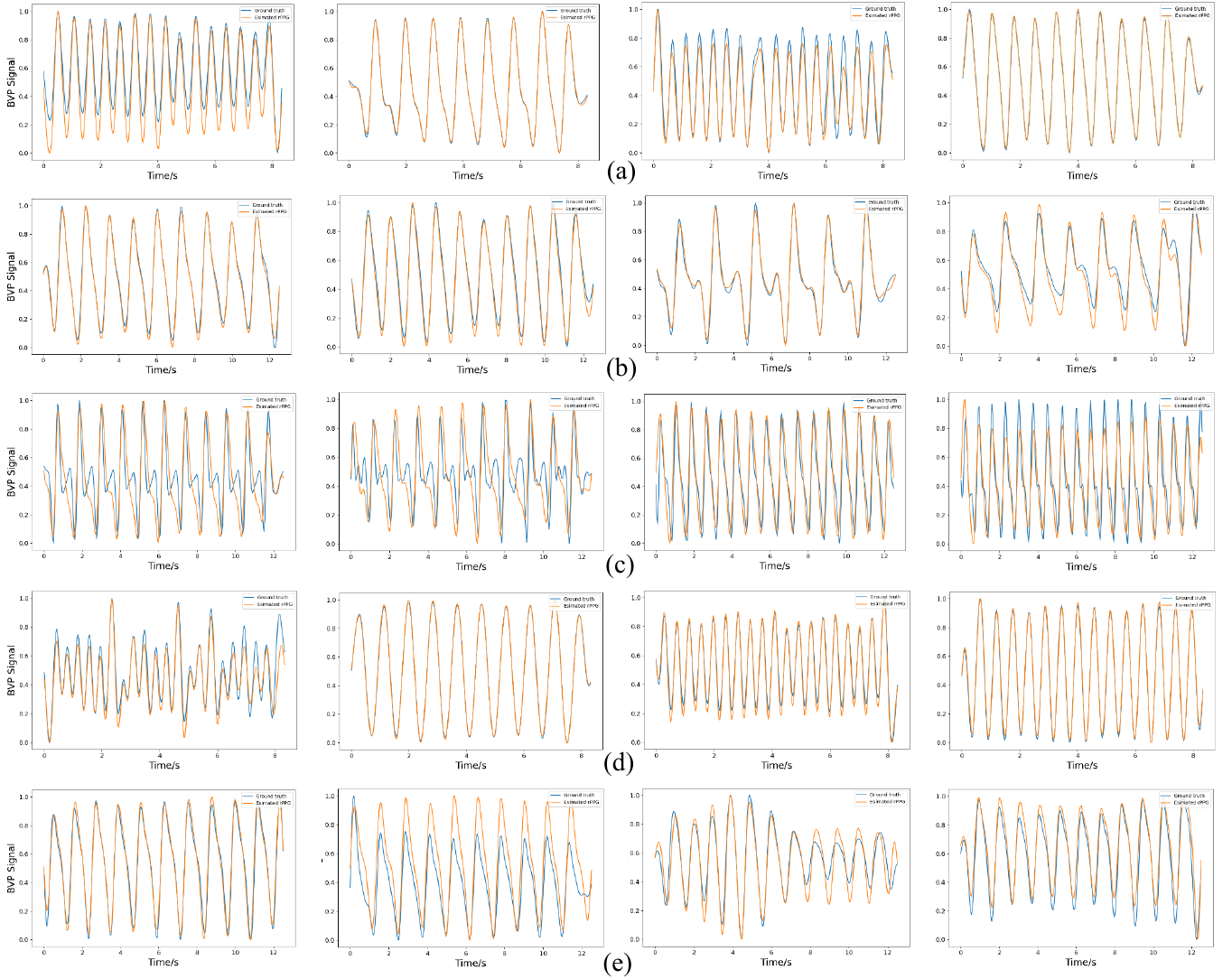}
	\caption{The visual comparison between estimated rPPG signals (orange curves) and their corresponding ground truth BVP signals (blue curves) on UBFC-rPPG, PURE, COHFACE, ECG-fitness and DEAP. The visual comparison of five datasets are corresponding to (a), (b), (c), (d) and (e).}
	\label{figure13}
\end{figure*}
Fig. \ref{figure13} depicts a comparative analysis between the estimated BVP signals generated by VidFormer and the ground truth BVP signals. The results highlight the effectiveness of VidFormer in precisely locating the peak systolic position of the cardiac cycle and reconstructing the BVP signal. Furthermore, Fig. \ref{figure13} represents the potential of VidFormer to reconstruct BVP signals across diverse environmental contexts.

\subsubsection{Cross-Dataset HR Evaluation}
To validate the generalization of the proposed method, we undertook cross-dataset testing across multiple benchmark datasets, namely UBFC, PURE and COHFACE. Subsequently, we juxtaposed the outcomes with those obtained from SOTA methodologies, as delineated in the Table \ref{table3}, Table \ref{table4} and Table \ref{table5}. For example, UBFC $\rightarrow$ PURE means training on UBFC-rPPG while testing on PURE. 
\begin{table}[h]
	\centering
	\caption{Comparison to State of the Art on Cross-Dataset HR Estimation}
	\label{table3}
	\begin{tabular}{p{60pt}cccccc}
	\toprule
		\multirow{2}{*}{\makecell[c]{Method}} &\multicolumn{3}{c}{UBFC$\rightarrow$PURE} &\multicolumn{3}{c}{PURE$\rightarrow$UBFC}\\
		\cmidrule(r){2-4} \cmidrule(r){5-7}
		~ &MAE &RMSE &$r$ &MAE &RMSE &$r$\\
		\midrule
		DeepPhys\cite{chen2018deepphys}$\blacktriangle$&$5.80$ &$9.10$ &$0.84$ &$5.02$ &$7.53$ &$0.64$\\
		PhysNet\cite{yu2019remote}$\blacktriangle$ &$8.39$ &$10.22$ &$0.71$ &$3.99$ &$5.49$ &$0.77$\\
		
		Meta-rppg\cite{lee2020meta}$\blacktriangle$ &$6.00$ &$12.98$ &$0.72$&$6.11$ &$7.58$ &$0.66$\\
		PulseGAN\cite{song2021pulsegan}$\star$ &$3.36$ &$5.11$ &$0.95$&$3.78$ &$5.12$ &$0.68$\\
		
		Physformer\cite{yu2022physformer}$\blacktriangle$ &$1.99$ &$3.28$ &$0.99$&$1.93$ &$3.02$ &$\mathbf{0.97}$\\
		Gideon\cite{gideon2021way}$\blacktriangle$ &$2.95$ &$4.60$ &$0.97$&$2.37$ &$3.51$ &$0.94$\\
		Yue\cite{yue2023facial}$\blacktriangle$ &$2.14$ &$3.37$ &$0.98$&$2.18$ &$3.20$ &$\mathbf{0.97}$\\
		Dual-GAN\cite{lu2021dual}$\star$ &$1.81$ &$\mathbf{2.97}$ &$0.99$&$2.03$ &$3.01$ &$\mathbf{0.97}$\\
		\textbf{VidFormer(ours)}$\blacktriangle$ &$\mathbf{1.79}$ &$3.21$ &$\mathbf{0.99}$&$\mathbf{1.92}$ &$\mathbf{2.99}$ &$\mathbf{0.97}$\\
	\bottomrule	
	\end{tabular}
\end{table}

\begin{table}[h]
	\centering
	\caption{Comparison to State of the Art on Cross-Dataset HR Estimation}
	\label{table4}
	\begin{tabular}{p{56pt}cccccc}
		\toprule
		\multirow{2}{*}{\makecell[c]{Method}} &\multicolumn{3}{c}{PURE$\rightarrow$COHFACE} &\multicolumn{3}{c}{COHFACE$\rightarrow$PURE}\\
		\cmidrule(r){2-4} \cmidrule(r){5-7}
		~ &MAE &RMSE &$r$ &MAE &RMSE &$r$\\
		\midrule
    	HRCNN\cite{vspetlik2018visual}$\blacktriangle$ &$-$ &$-$ &$-$ &$8.72$ &$11.0$ &$0.70$\\
		Two-stream CNN\cite{wang2019vision}$\star$ &$-$ &$-$ &$-$ &$9.81$ &$11.81$ &$0.42$\\
		
		DeeprPPG\cite{liu2020general}$\blacktriangle$ &$7.66$ &$13.35$ &$0.46$&$6.55$ &$20.83$ &$0.54$\\
		\textbf{VidFormer(ours)}$\blacktriangle$ &$\mathbf{4.79}$ &$\mathbf{7.21}$ &$\mathbf{0.85}$&$\mathbf{2.21}$ &$\mathbf{4.03}$ &$\mathbf{0.91}$\\
		\bottomrule	
	\end{tabular}
\end{table}

\begin{table}[t]
	\centering
	\caption{Comparison to State of the Art on Cross-Dataset HR Estimation}
	\label{table5}
	\begin{tabular}{p{56pt}cccccc}
		\toprule
		\multirow{2}{*}{\makecell[c]{Method}} &\multicolumn{3}{c}{UBFC$\rightarrow$COHFACE} &\multicolumn{3}{c}{COHFACE$\rightarrow$UBFC}\\
		\cmidrule(r){2-4} \cmidrule(r){5-7}
		~ &MAE &RMSE &$r$ &MAE &RMSE &$r$\\
		\midrule		
		DeeprPPG\cite{liu2020general}$\blacktriangle$ &$4.05$ &$10.66$ &$0.70$&$4.52$ &$9.69$ &$0.86$\\
		\textbf{VidFormer(ours)}$\blacktriangle$ &$\mathbf{3.78}$ &$\mathbf{8.32}$ &$\mathbf{0.89}$ &$\mathbf{3.23}$ &$\mathbf{6.19}$ &$\mathbf{0.91}$\\
		\bottomrule	
	\end{tabular}
\end{table}
The tabulated data illustrates that our proposed method demonstrates commendable performance, exhibiting competitiveness when juxtaposed with existing methods. Notably, in some cross-dataset testing experiments, our method has attained parity with  SOTA benchmarks. Such observations suggest the robust generalization capabilities inherent within our proposed approach.

\section{Ablation Experiments}
We conduct ablation study from five aspects: GA-3DCNN, ST-MHSA, Local Convolution Branch, Global Transformer Branch and CTIM. The results for HR estimation are reported on PURE and DEAP datasets. 

\subsection{GA-3DCNN}
The GA-3DCNN model is specifically engineered to discern and extract intricate local spatial and temporal features within the input data. Consequently, in order to ascertain the efficacy of the proposed attention mechanism, we tested it on the PURE and DEAP datasets without adding temporal attention of global attention (T-GA), spatial attention of global attention (S-GA), and global attention (GA) as shown in Table \ref{table6}.
\begin{table}[h]
	\centering
	\caption{Ablation Study on the GA-3DCNN}
	\label{table6}
	\begin{tabular}{ccccccc}
		\toprule
		\multirow{2}{*}{\makecell[c]{Method}} &\multicolumn{3}{c}{PURE} &\multicolumn{3}{c}{DEAP}\\
		\cmidrule(r){2-4} \cmidrule(r){5-7}
		~ &MAE &RMSE &$r$ &MAE &RMSE &$r$\\
		\midrule		
		Without T-GA &$1.25$ &$4.12$ &$0.97$&$2.75$ &$5.71$ &$0.93$\\
		Without S-GA &$1.28$ &$4.99$ &$0.97$ &$2.18$ &$5.44$ &$0.94$\\
		Without GA &$1.87$ &$5.72$ &$0.95$&$3.04$ &$6.69$ &$0.93$\\
		VidFormer &$0.42$ &$1.18$ &$0.99$ &$0.75$ &$1.46$ &$0.98$\\
		\bottomrule	
	\end{tabular}
\end{table}

Table \ref{table6} demonstrates that the removal of the GA module results in a noticeable decline in the reconstruction of BVP signal. Notably, the error associated with the removal of T-GA is less significant than that observed with the removal of S-GA. This discrepancy arises because the reconstruction of the BVP signal involves identifying a potential mapping relationship between the BVP signal and skin color and minimizing the error between this mapping and the true relationship. Consequently, exploring the color distribution relationship across various spatial regions becomes crucial, highlighting the significance of the S-GA module. On the other hand, we expect T-GA to assist 3DCNN in capturing possible time-related signals, such as environmental noise, lighting changes, personnel movement, etc., so that the Local Convolution Branch can attempt to decouple these signals. Therefore, when eliminating T-GA, the reconstruction accuracy of BVP signals also decreases.

\subsection{ST-MHSA}
Our aim is to formulate a Global Transformer module tailored for capturing the nuanced dynamics of head movement and lighting alterations. This necessitates adept feature extraction from temporal as well as spatial domains. Consequently, in order to assess the effectiveness of the designed attention mechanism, we performed ablation studies specifically targeting the ST-MHSA module, including Spatial-MHSA (S-MHSA), Time-MHSA (T-MHSA) as indicated in Table \ref{table7}.
\begin{table}[h]
	\centering
	\caption{Ablation Study on the ST-MHSA}
	\label{table7}
	\begin{tabular}{ccccccc}
		\toprule
		\multirow{2}{*}{\makecell[c]{Method}} &\multicolumn{3}{c}{PURE} &\multicolumn{3}{c}{DEAP}\\
		\cmidrule(r){2-4} \cmidrule(r){5-7}
		~ &MAE &RMSE &$r$ &MAE &RMSE &$r$\\
		\midrule		
		Without S-MHSA &$2.36$ &$6.52$ &$0.96$&$4.74$ &$6.66$ &$0.90$\\
		Without T-MHSA &$0.79$ &$3.42$ &$0.98$ &$1.21$ &$2.52$ &$0.96$\\
		VidFormer &$0.42$ &$1.18$ &$0.99$ &$0.75$ &$1.46$ &$0.98$\\
		\bottomrule	
	\end{tabular}
\end{table}

In Table \ref{table7}, it is noteworthy that the removal of S-MHSA and T-MHSA components within the ST-MHSA framework leads to an increase in both MAE and RMSE in the predicted HR, accompanied by a decrease in the $r$. This phenomenon can be attributed to the susceptibility of human facial skin color in the input video to various factors such as head movements and fluctuations in lighting conditions. Effective extraction of Blood Volume Pulse (BVP) signals necessitates discerning the spatial relationships between different skin blocks and background elements, as well as comprehending temporal variations in skin color. Consequently, feature extraction from both spatial and temporal dimensions of the input video is imperative. Hence, the omission of either S-MHSA or T-MHSA components results in diminished accuracy in reconstructing the BVP signal.

However, it is important to highlight that the exclusion of S-MHSA, as opposed to the elimination of T-MHSA, results in a more pronounced reduction in the accuracy of BVP signal reconstruction. This observation underscores the important contributions of spatial features to the final reconstruction of the BVP signal compare with temporal features. The correlation between the ground truth BVP signal and the RGB image captured by the camera is notable. When the sampling frequency of the BVP signal aligns with the video frame rate, it implies that each individual frame of the image corresponds to a specific temporal point within the BVP signal, as illustrated in (\ref{equation01}) and Fig. \ref{figure0}. Consequently, employing neural networks to fit the mapping from BVP signals to RGB images enhances the focus of network on the spatial characteristics inherent in the input video data. Nevertheless, the dynamic movement of head of the subject in the video footage may induce variations in the pixel coordinates corresponding to the skin region across successive frames. Hence, the integration of temporal attention mechanisms becomes imperative, as they facilitate the establishment of temporal correlations among disparate frames.

\subsection{Local Convolution Branch}
The Local Convolution Branch is designed to extract local information from input data and model head movements within a specific time window, thereby highlighting areas requiring special attention. Owing to the inductive bias of locality and translation invariance inherent in CNNs, these networks can reconstruct BVP signals effectively even when the head of an individual experiences short-distance translations and slight rotational movements. Furthermore, due to this inductive bias, CNNs often perform better on smaller datasets compared to Transformers. Based on the above analysis, we conducted ablation experiments on the Local Convolution Branch.

First, we set the same parameters and experimental environment to evaluate the performance of models with the Local Convolution Branch (w-LCB) and without the Local Convolution Branch (wo-LCB) on the PURE and DEAP datasets as indicated in Table \ref{table8}.
\begin{table}[h]
	\centering
	\caption{Ablation Study on the Local Convolution Branch}
	\label{table8}
	\begin{tabular}{ccccccc}
		\toprule
		\multirow{2}{*}{\makecell[c]{Method}} &\multicolumn{3}{c}{PURE} &\multicolumn{3}{c}{DEAP}\\
		\cmidrule(r){2-4} \cmidrule(r){5-7}
		~ &MAE &RMSE &$r$ &MAE &RMSE &$r$\\
		\midrule		
		wo-LCB &$36.97$ &$62.83$ &$0.12$ &$24.84$ &$28.67$ &$0.44$\\
		w-LCB  &$0.42$ &$1.18$ &$0.99$ &$0.75$ &$1.46$ &$0.98$\\
		\bottomrule	
	\end{tabular}
\end{table}

\begin{figure}[h]
	\centering
	\includegraphics[scale=0.32]{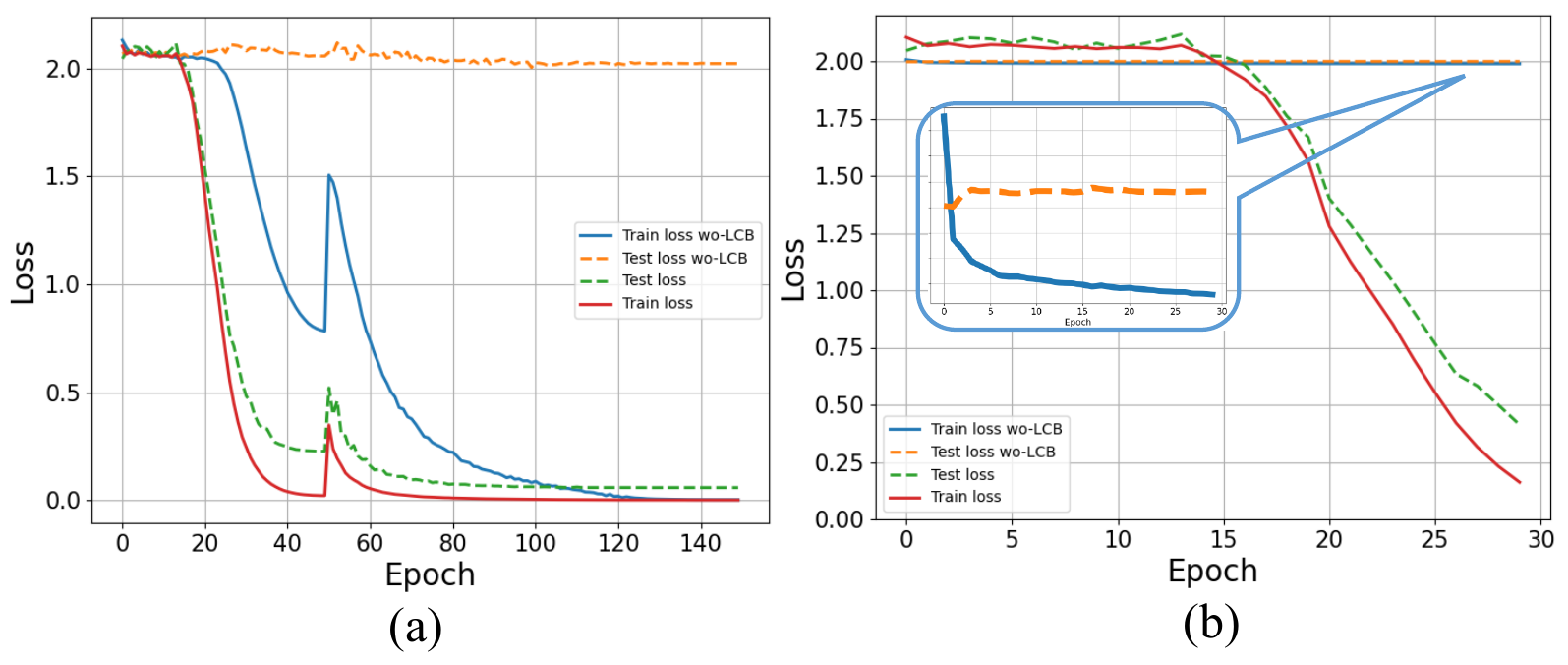}
	\caption{The training and testing loss curves of wo-LCB and w-LCB on PURE and DEAP datasets. (a) presents the training and testing loss on PURE dataset and (b) indicates the training and testing loss on DEAP dataset.}
	\label{figure14}
\end{figure}

From Table \ref{table8}, it can be observed that when the Local Convolution Branch is eliminated and only the Transformer branch is present in the model. The accuracy of BVP signal reconstruction has significantly decreased. Meanwhile, we will compare the convergence curves of VidFormer with VidFormer that eliminates Local Convolution Branch as shown in Fig. \ref{figure14}.
Fig. \ref{figure14} illustrates the loss curves for w-LCB and wo-LCB on the PURE and DEAP datasets. Notably, on the PURE dataset, the wo-LCB model effectively fits the training set but struggles with the test set, resulting in a MAE of 36.97 bpm in HR estimation for the test set. Additionally, the w-LCB model demonstrates faster convergence on the PURE dataset compared to the wo-LCB model. Another significant observation is that, given the same number of iterations, the wo-LCB model exhibits more difficulty in converging on both the training and testing sets of the DEAP dataset compared to the w-LCB model. Consequently, this analysis suggests that incorporating a Local Convolution Branch introduces an inductive bias to the network, thereby enhancing its fitting ability.

\subsection{Global Transformer Branch}
\begin{figure}[h]
	\centering
	\includegraphics[scale=0.31]{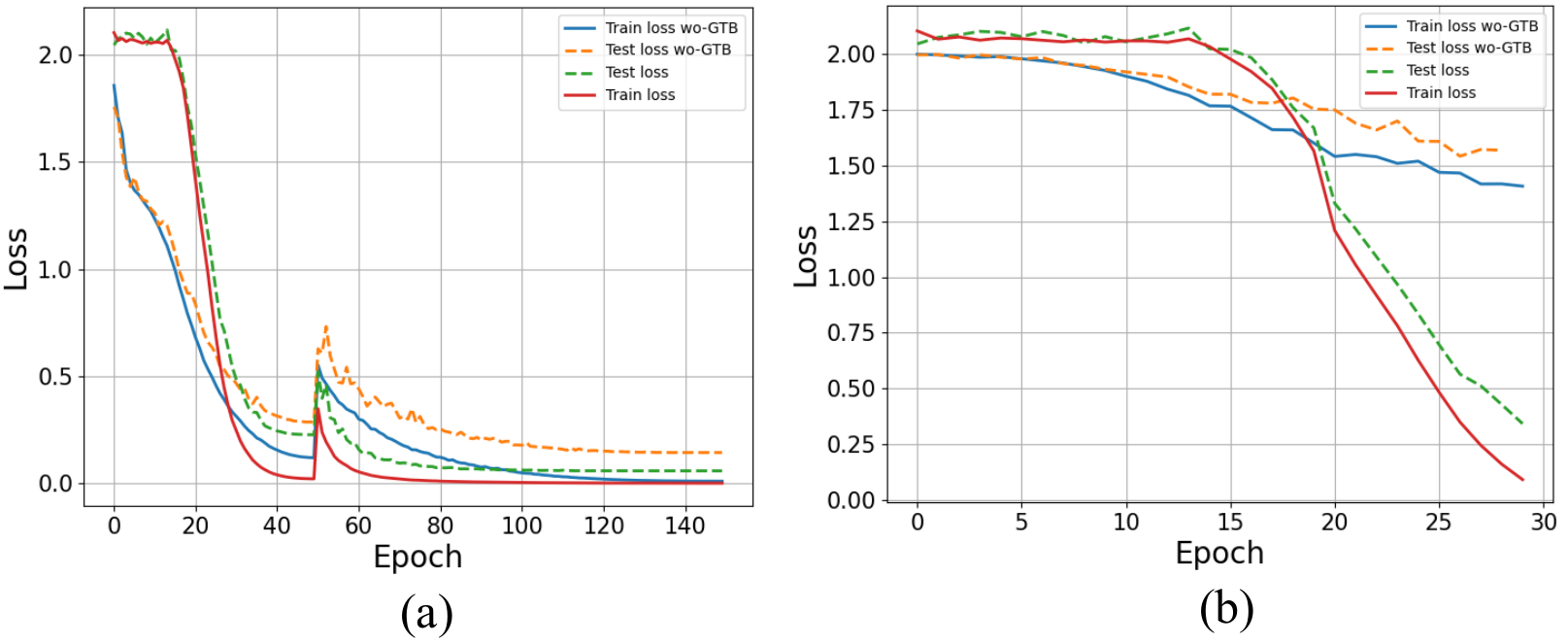}
	\caption{The training and testing loss curves of wo-GTB and w-GTB on PURE and DEAP datasets. (a) presents the training and testing loss on PURE dataset and (b) indicates the training and testing loss on DEAP dataset.}
	\label{figure15}
\end{figure}

Based on the analysis in subsection \ref{GTB}, we design the Global Transformer Branch to alleviate the lack of global modeling capability in the Local Revolution Branch. Due to the use of small convolution kernels and shallow network layers in the Local Convolution Branch, it is difficult to effectively model the input data globally. Therefore, in this chapter, we conducted ablation experiments on the Global Transformer Branch. Firstly, we set up the same experimental environment for models with Global Transformer Branch (w-GTB) and without Global Transformer Branch (wo-GTB), and then trained and tested them on the PURE and DEAP datasets, and the HR estimation results are shown in Table \ref{table9}.
\begin{table}[h]
	\centering
	\caption{Ablation Study on the Global Transformer Branch}
	\label{table9}
	\begin{tabular}{ccccccc}
		\toprule
		\multirow{2}{*}{\makecell[c]{Method}} &\multicolumn{3}{c}{PURE} &\multicolumn{3}{c}{DEAP}\\
		\cmidrule(r){2-4} \cmidrule(r){5-7}
		~ &MAE &RMSE &$r$ &MAE &RMSE &$r$\\
		\midrule		
		wo-GTB &$3.87$ &$34.53$ &$0.82$ &$52.31$ &$67.78$ &$0.03$\\
		w-GTB  &$0.42$ &$1.18$ &$0.99$ &$0.75$ &$1.46$ &$0.98$\\
		\bottomrule	
	\end{tabular}
\end{table}
Table \ref{table9} reveals that the wo-GTB can estimate the HR on the PURE dataset and achieve a low MAE. However, the RMSE for wo-GTB is significantly high, indicating that the HR estimation error for some samples is huge. This is likely due to the inductive bias of CNNs\cite{liu2024transformer}, which causes it to favor some data patterns. Additionally, the HR estimation results of wo-GTB on the DEAP dataset show significant errors. This is attributed to the large size and higher complexity of the DEAP dataset, making it challenging for wo-GTB to converge within a limited number of iterations. Furthermore, a comparison of the loss curves of wo-GTB and w-GTB is presented in Fig. \ref{figure15}.

From Fig. \ref{figure15}, it can be observed that although wo-GTB and w-GTB can converge on the PURE dataset, w-GTB converges faster than wo-GTB and has lower losses on the test set, resulting in better HR estimation results.
Another noteworthy point is that wo-GTB struggles to converge on the DEAP dataset. We attribute this to two primary reasons. First, the complexity and the size of DEAP dataset make it appear some data less suitable for CNNs, thereby hindering convergence. Second, compared to CNNs, Transformers offer more effective global modeling, which enhances performance on large datasets. 

\subsection{CTIM}
Based on the results of the aforementioned ablation experiments, both the Local Convolution Branch and the Global Transformer Branch are essential components of VidFormer. The interaction between these two branches is equally critical. Consequently, it is necessary to design ablation experiments for the CTIM module including Conv-Trans Block (C-TB) and Trans-Conv Block (T-CB) to verify the effectiveness of information exchange between the two branches as shown in Table \ref{table10}.
\begin{table}[t]
	\centering
	\caption{Ablation Study on the ST-MHSA}
	\label{table10}
	\begin{tabular}{ccccccc}
		\toprule
		\multirow{2}{*}{\makecell[c]{Method}} &\multicolumn{3}{c}{PURE} &\multicolumn{3}{c}{DEAP}\\
		\cmidrule(r){2-4} \cmidrule(r){5-7}
		~ &MAE &RMSE &$r$ &MAE &RMSE &$r$\\
		\midrule		
		Without C-TB &$10.87$ &$16.12$ &$0.62$&$18.94$ &$25.67$ &$0.54$\\
		Without T-CB &$2.56$ &$8.09$ &$0.97$ &$5.21$ &$10.39$ &$0.90$\\
		VidFormer &$0.42$ &$1.18$ &$0.99$ &$0.75$ &$1.46$ &$0.98$\\
		\bottomrule	
	\end{tabular}
\end{table}

Table \ref{table10} demonstrates that eliminating the C-TB results in significant errors in HR estimation, similar to the errors observed when eliminating the Local Convolution Branch. This issue arises because the Transformer lacks inductive bias, making it challenging to effectively fit small datasets. Conversely, the presence of the C-TB introduces inductive bias to the Global Transformer Branch, thereby enhancing the accuracy of HR estimation. Furthermore, eliminating the T-CB also reduces the accuracy of HR estimation. This is because the information extracted by the Global Transformer Branch effectively enhances the global modeling capability of the Local Convolution Branch, thereby aiding in the extraction of more generalized representation information. Therefore, CTIM can effectively exchange and fuse information from two branches.

\section{Discussion}
In this section, we discuss the impact of ethnicity, human exercise, and makeup on our model separately.
\subsection{Effect of Ethnicity}
First, we examine the impact of different racial groups on the reconstruction of BVP signals using our proposed model. To validate this, we utilized the COHFACE dataset, which encompasses a diverse range of ethnicities including Caucasian, African, and Asian (both East Asian and Middle Eastern). We divided the COHFACE dataset into three subsets based on race: COHFACE-Asian, COHFACE-Caucasian, and COHFACE-African. Our proposed model was then trained and tested separately on each of these three subsets to evaluate its performance across different racial groups. HR estimation results of the three datasets can be seen in the Table \ref{table11}.
\begin{table*}[t]
	\centering
	\caption{HR Estimation for Subjects of Different Ethnicities}
	\label{table11}
	\begin{tabular}{cccccccccc}
		\toprule
		\multirow{2}{*}{\makecell[c]{Method}} &\multicolumn{3}{c}{COHFACE-Caucasian} &\multicolumn{3}{c}{COHFACE-African} &\multicolumn{3}{c}{COHFACE-Asian}\\
		\cmidrule(r){2-4} \cmidrule(r){5-7} \cmidrule{8-10}
		~ &MAE &RMSE &$r$ &MAE &RMSE &$r$ &MAE &RMSE &$r$\\
		\midrule		
		PFE-TFA\cite{li2023learning}$\blacktriangle$ &$1.13$ &$3.61$ &$0.97$&$1.67$ &$4.02$ &$0.94$ &$1.21$ &$3.54$ &$0.97$\\
		CPulse\cite{mehta2023cpulse}$\star$ &$1.59$ &$3.42$ &$0.94$ &$2.52$ &$4.87$ &$0.90$ &$1.48$ &$3.70$ &$0.94$\\
		\textbf{VidFormer(ours)}$\blacktriangle$ &$0.51$ &$1.06$ &$0.99$ &$0.69$ &$1.66$ &$0.98$ &$0.47$ &$0.98$ &$0.99$\\
		\bottomrule	
	\end{tabular}
\end{table*}

\begin{figure}[h]
	\centering
	\includegraphics[scale=0.32]{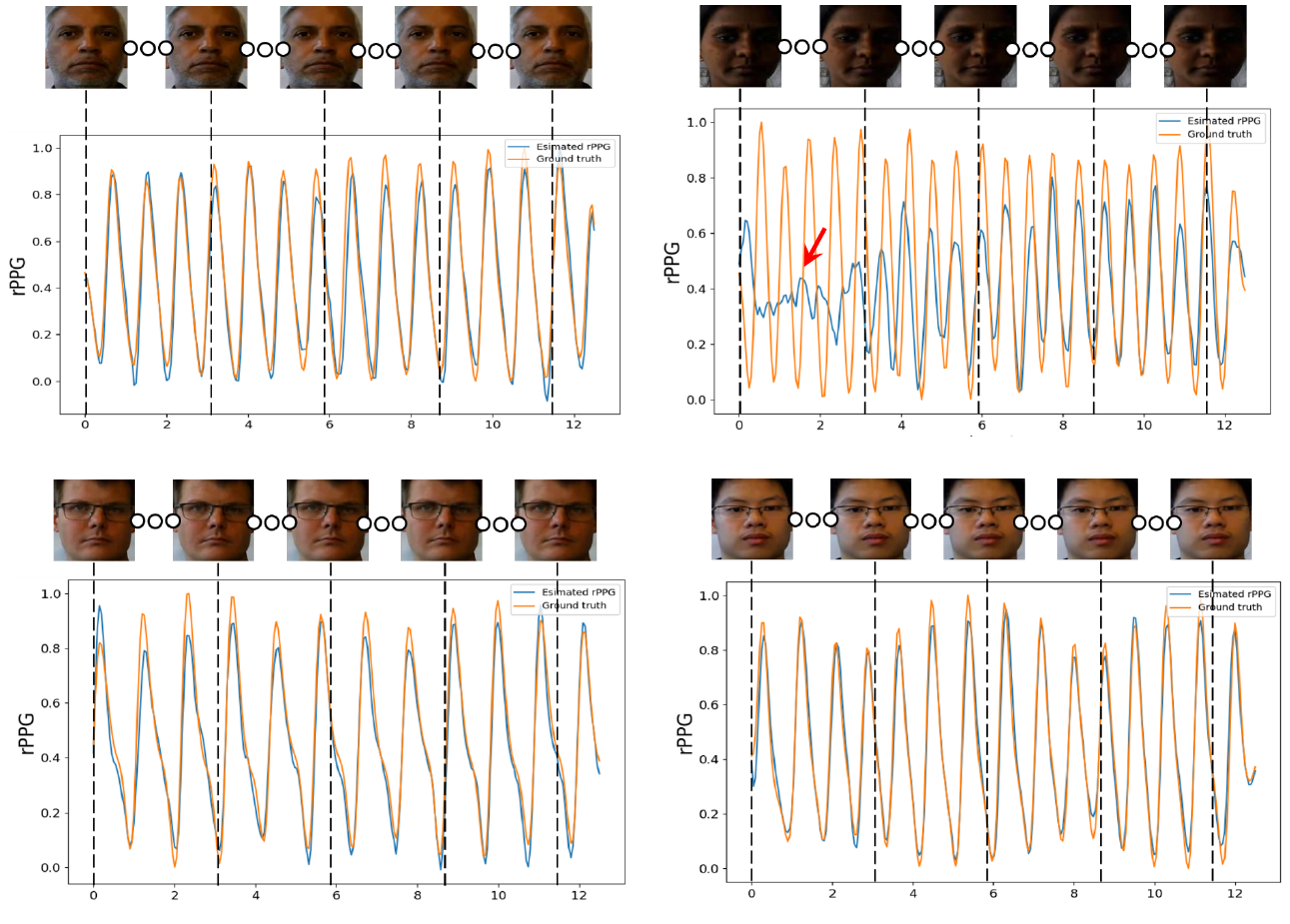}
	\caption{The performance of BVP signal reconstruction on subjects of different ethnicity.}
	\label{figure16}
\end{figure}

Table \ref{table11} indicates that our proposed method performs best on the COHFACE-Asian, and also achieve the best results on both the COHFACE-Caucasian and COHFACE-African. Notably, the HR estimation results for COHFACE African exhibit the largest error. This discrepancy arises because most RGB cameras are optimized to capture light skin tones more effectively than dark skin tones. Typically, the sensitivity of camera peaks in the middle of the (0, 255) pixel range, meaning that the darker skin tones of African subjects are more likely to saturate the pixels. This saturation suppresses the skin color changes caused by physiological signal variations\cite{mcduff2023camera}. In addition, we present examples of BVP signal reconstruction for the subjects of different ethnicity in Fig. \ref{figure16}.

\subsection{Effect of Makeup}
Another challenge in rPPG tasks is the presence of makeup on testers. rPPG estimates HR changes based on the absorption and reflection of light, which are influenced by blood flow volumn in the blood vessels. When a tester wears makeup, it can alter the absorption and reflection intensity of light, introducing additional environmental noise and potentially compromising the accuracy of HR estimations. Therefore, we categorized the COHFACE dataset into two subsets based on the presence of makeup: COHFACE without makeup (COHFACE-woM) and COHFACE with makeup (COHFACE-wM). Each subset was further divided into training and testing sets. Separate training and testing were then conducted on their respective training and testing sets, the HR estimation results can be seen in the Table \ref{table12}.
\begin{table}[h]
	\centering
	\caption{HR Estimation for Subjects of Different Ethnicities}
	\label{table12}
	\begin{tabular}{ccccccc}
		\toprule
		\multirow{2}{*}{\makecell[c]{Method}} &\multicolumn{3}{c}{COHFACE-woM} &\multicolumn{3}{c}{COHFACE-African-wM}\\
		\cmidrule(r){2-4} \cmidrule(r){5-7}
		~ &MAE &RMSE &$r$ &MAE &RMSE &$r$\\
		\midrule		
		PFE-TFA\cite{li2023learning}$\blacktriangle$ &$1.28$ &$3.78$ &$0.96$&$1.37$ &$3.93$ &$0.95$ \\
		CPulse\cite{mehta2023cpulse}$\star$ &$1.99$ &$3.76$ &$0.94$ &$2.07$ &$4.94$ &$0.92$\\
		\textbf{VidFormer(ours)}$\blacktriangle$ &$0.48$ &$1.36$ &$0.99$ &$0.58$ &$0.84$ &$0.99$ \\
		\bottomrule	
	\end{tabular}
\end{table}

\begin{figure}[h]
	\centering
	\includegraphics[scale=0.32]{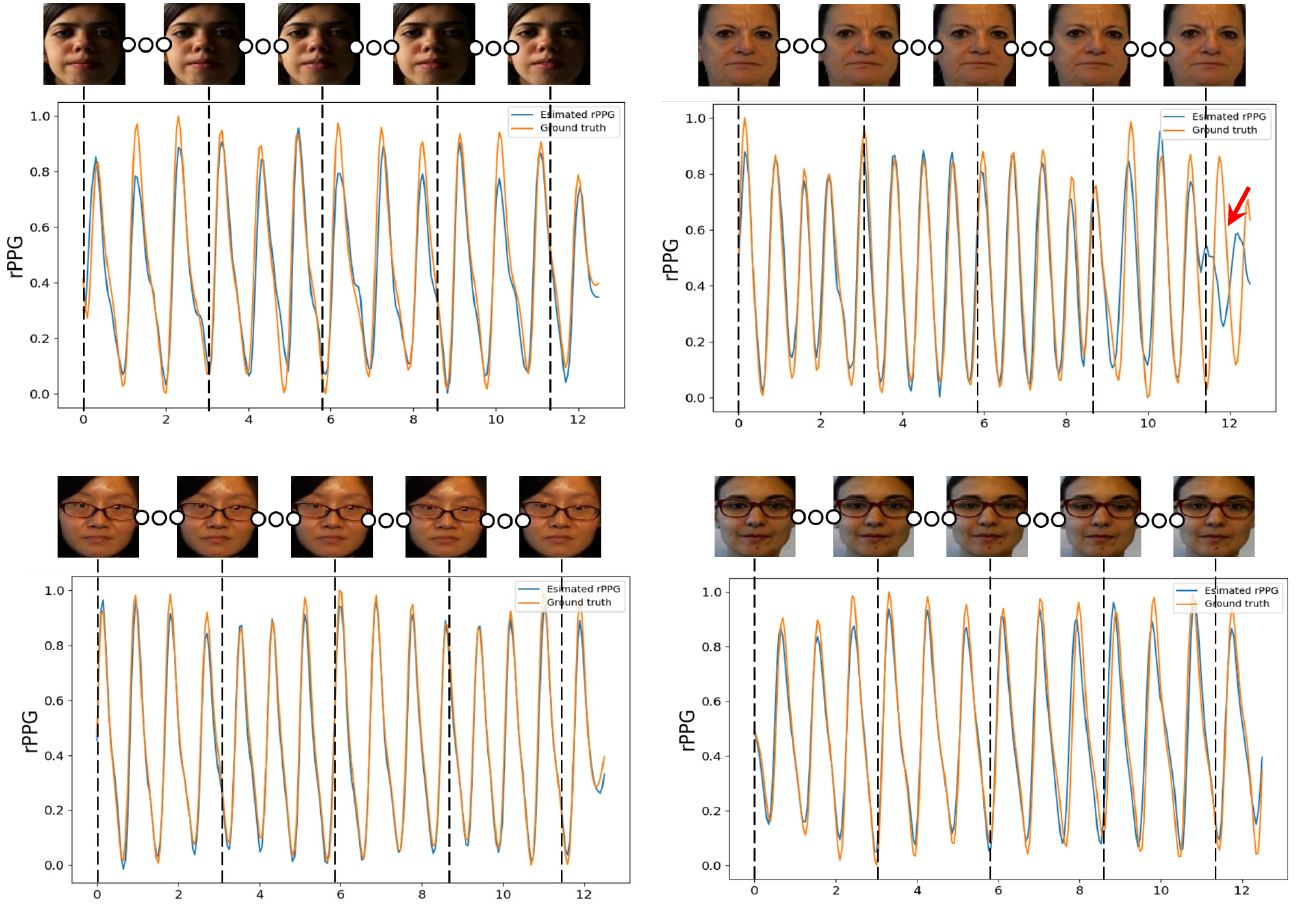}
	\caption{The performance of BVP signal reconstruction on subjects with makeup.}
	\label{figure17}
\end{figure}

Table \ref{table12} reveals that while all methods exhibit a decline in HR estimation accuracy when makeup is applied on subjects, our proposed method remains highly competitive. This stability is due to the consistent absorption and reflection of light by makeup, which remains within a stable range when the degree of makeup is fixed\cite{yoshida2021estimation}. The T-MHSA and T-GA in VidFormer consider input data over the temporal dimension, enabling VidFormer to effectively capture the temporal transformations of BVP signals. Furthermore, the BVP signal reconstruction results for different subjects with makeup can be seen in the Fig. \ref{figure17}.

\subsection{Effect of Exercise}
\begin{figure}[h]
	\centering
	\includegraphics[scale=0.32]{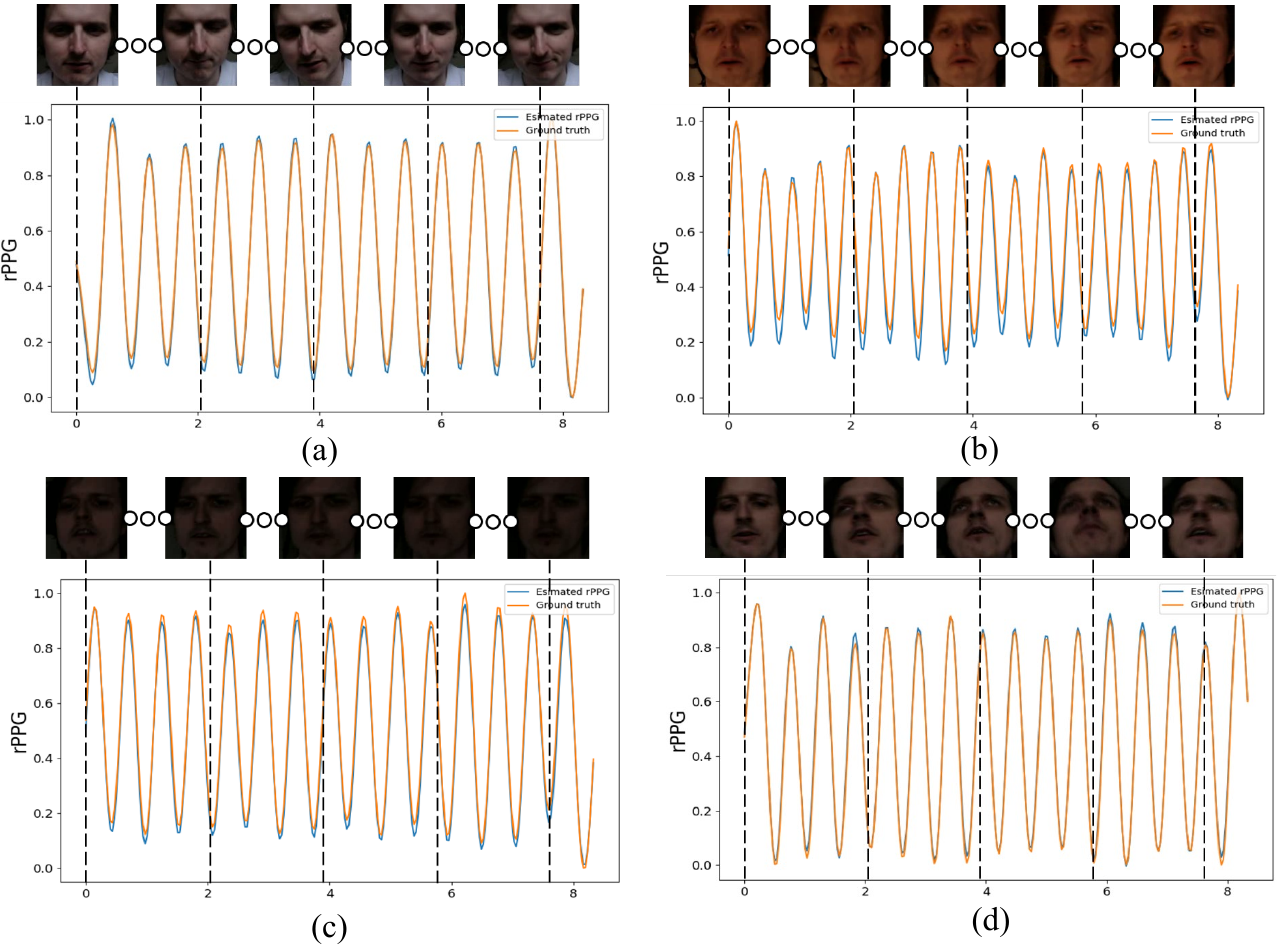}
	\caption{The performance of BVP signal reconstruction on subjects with exercise. (a) shows the elliptical machine used by testers under natural light, (b) is the rowing machine used by testers under artificial light, (c) indicates the rowing machine used by testers under natural light and (d) shows the stationary bike movements used by testers under natural light}
	\label{figure18}
\end{figure}

In the final chapter, we will examine the impact of tester movement on the performance of VidFormer. The ECG-fitness dataset includes data from testers using rowing machines, elliptical trainers and stationary
bike movements, which introduces substantial facial movements and significant variations in lighting intensity. We trained and tested VidFormer using this dataset, and the results of BVP signal reconstruction are presented in Fig. \ref{figure18}.

Fig. \ref{figure18} reveals that VidFormer can still achieve good BVP signal reconstruction results even when testing personnel are in motion. The possible reason is that ST-MHSA and GA-3DCNN in VidFormer can effectively extract spatiotemporal features of input data to provide model robustness.

\section{Conclusion}
In this paper, we introduce VidFormer, which comprises five key modules: Stem, Local Convolution Branch, Global Transformer Branch, CTIM, and the RGM. We propose an improved dichromatic model tailored to BVP signal reconstruction and VidFormer is designed based on this enhanced model. The Stem module is responsible for extracting preliminary features from the input data. The Local Convolution Branch captures local features and provides prior information to the Global Transformer Branch. Local Convolution Branch includes GA-3DCNN and BS-3DCNN, where GA contributes global spatiotemporal features, mitigating data bias. The Global Transformer Branch extracts global features from the input data. It includes S-MHSA, which establishes relationships between different skin regions, and T-MHSA, which models movement and lighting changes over time. The CTIM module facilitates the exchange and fusion of features between the Local Convolution Branch and the Global Transformer Branch, reducing feature biases and introducing inductive bias to the Global Transformer Branch. The RGM is designed to produce BVP signals. For network optimization, we employ L1 loss and the negative Pearson correlation coefficient as optimization functions. VidFormer was evaluated on five datasets, demonstrating superior performance compared to the state-of-the-art methods. Finally, we discuss the essential roles of each VidFormer module and examine the effects of ethnicity, makeup, and Exercise on its performance.

\bibliography{IEEEabrv,references}
\bibliographystyle{IEEEtran}

\vspace{-28pt}
\begin{IEEEbiography}  [{\includegraphics[width=1in,height=1.25in,clip,keepaspectratio]{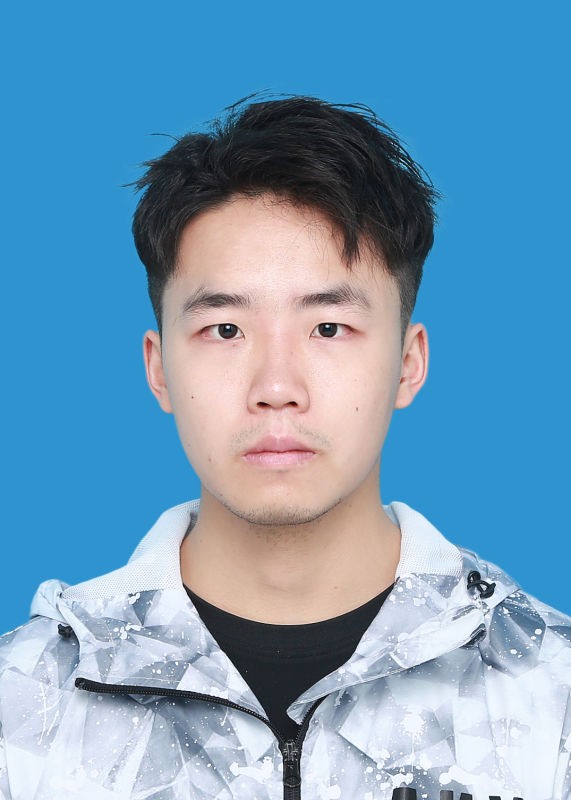}}] 
	{Jiachen Li} received the B.S. degree in information engineering from the Central South University, ChangSha, China, in 2022. He is currently pursuing the M.S. degree with the School of Information and Communication Engineering from the University of Electronic Science and Technology of China, Chengdu, China. His research interests include millimeter wave radar, human vital sign detection and deep learning.
\end{IEEEbiography} 
\vspace{-28pt}
\begin{IEEEbiography}  [{\includegraphics[width=1in,height=1.25in,clip,keepaspectratio]{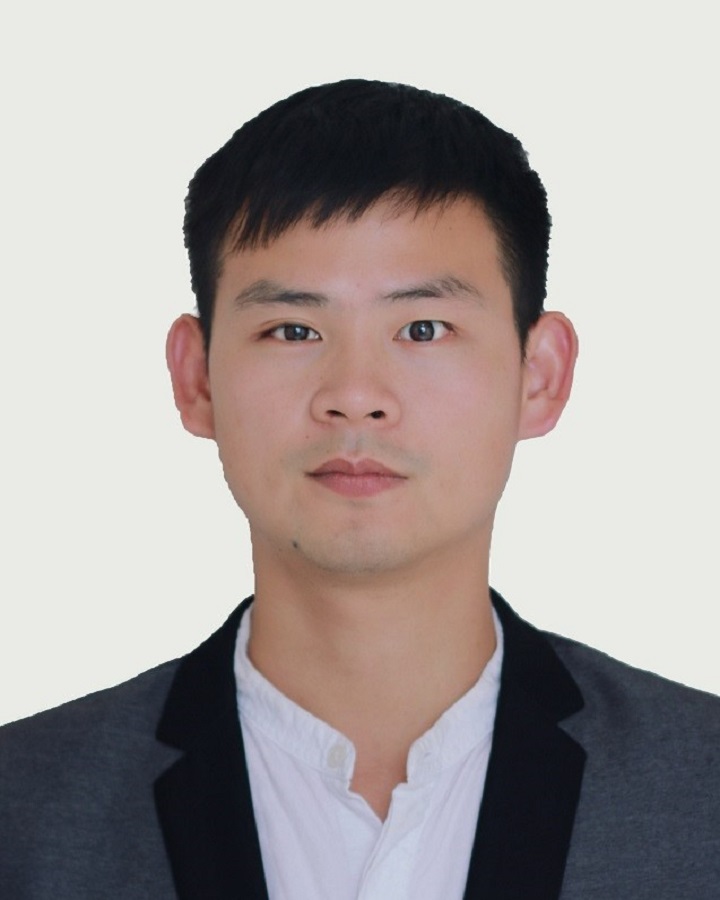}}] 
	{Shisheng Guo} (Member, IEEE) received the B.S. degree in communication engineering from the Nanchang Hangkong University, Nanchang, China, in 2013, and the Ph.D. degree in signal and information processing from the University of Electronic Science and Technology of China (UESTC), Chengdu, China, in 2019. He is currently an Associate Researcher with the School of Information and Communication Engineering, UESTC. His research interests include through-the-wall radar imaging, signal analysis and NLOS target detection.
\end{IEEEbiography} 
\vspace{-28pt}
\begin{IEEEbiography}  [{\includegraphics[width=1in,height=1.25in,clip,keepaspectratio]{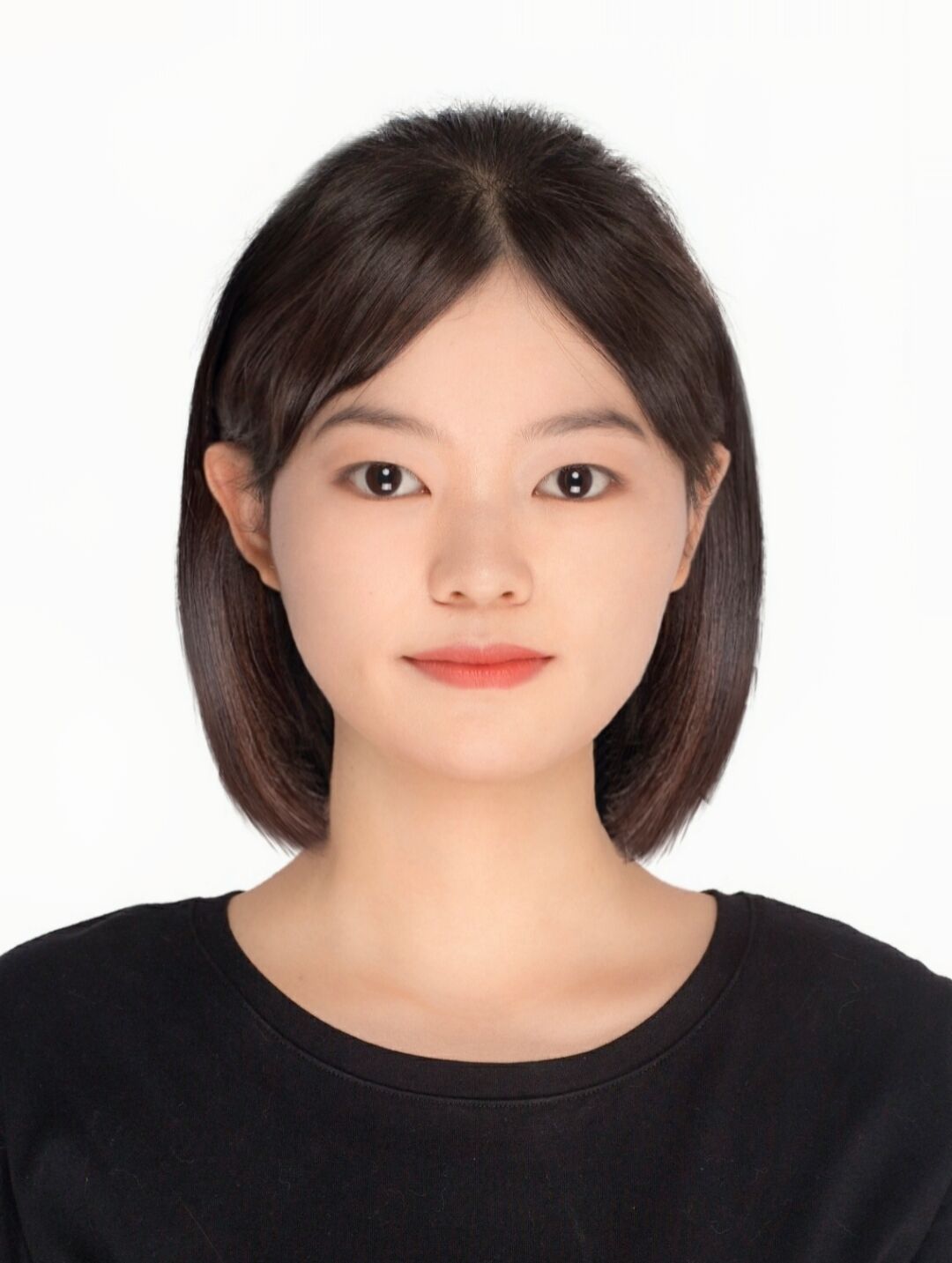}}] 
	{Longzhen Tang}  received the B.S. degree in information engineering from the Chengdu University
	of Technology, Chengdu, China, in 2022. She is currently pursuing the M.S. degree with the School
	of Information and Communication Engineering, University of Electronic Science and Technology of
	China, Chengdu. Her research interests include ultrawideband radar, human activity recognition, and deep learning.
\end{IEEEbiography} 
\vspace{-28pt}
\begin{IEEEbiography}  [{\includegraphics[width=1in,height=1.25in,clip,keepaspectratio]{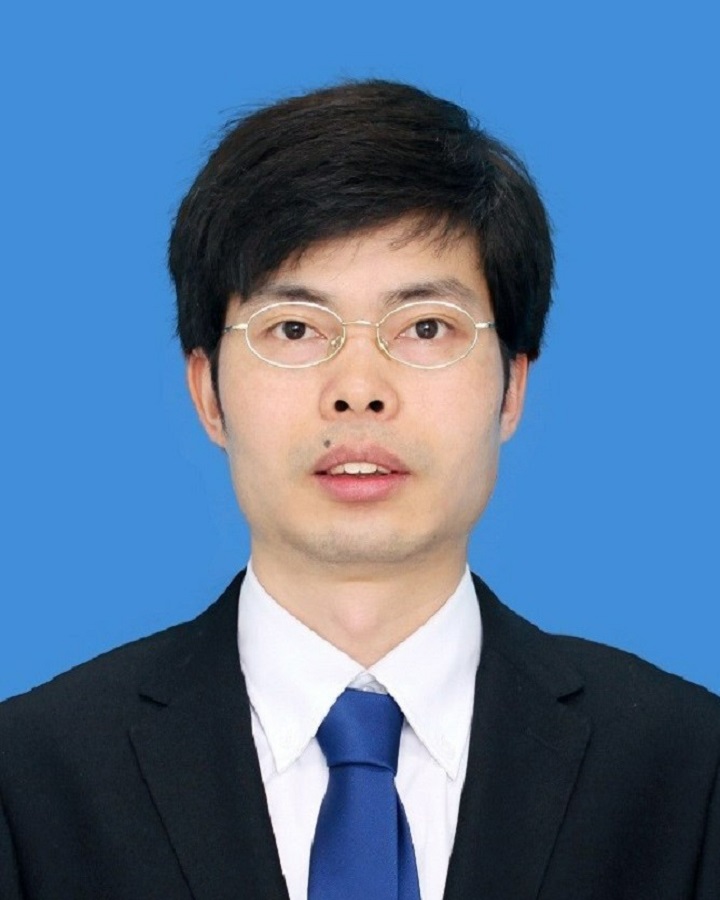}}] 
	{Guolong Cui} (Senior Member, IEEE) received the B.S. degree in electronic information engineering, M.S. and Ph.D. degrees in signal and information processing from the University of Electronic Science and Technology of China (UESTC), Chengdu, China, in 2005, 2008, and 2012, respectively. From January 2011 to April 2011, he was a Visiting Researcher with the University of Naples Federico II, Naples, Italy. From June 2012 to August 2013, he was a Postdoctoral Researcher with the Department of Electrical and Computer Engineering, Stevens Institute of Technology, Hoboken, NJ, USA. From September 2013 to July 2018, he was an Associate Professor in UESTC, where since August 2018, he has been a Professor. His current research interests include cognitive radar, array signal processing, MIMO radar, and through-the-wall radar.
\end{IEEEbiography} 
\vspace{-28pt}
\begin{IEEEbiography}  [{\includegraphics[width=1in,height=1.25in,clip,keepaspectratio]{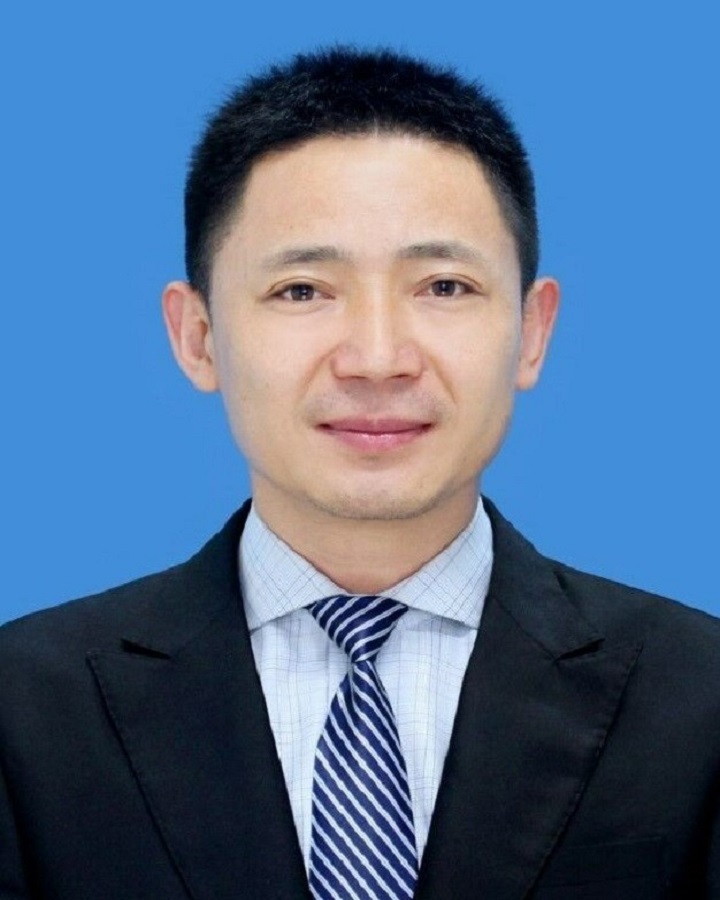}}] 
	{Lingjiang Kong} (Senior Member, IEEE) was born in 1974. He received the B.S. degree in electronic engineering, and M.S. and Ph.D. degrees in signal and information processing from the University of Electronic Science and Technology of China, Chengdu, China, in 1997, 2000, and 2003, respectively. From September 2009 to March 2010, he was a Visiting Researcher with the University of Florida. He is currently a Professor with the School of Information and Communication Engineering, University of Electronic Science and Technology of China. His research interests include Multiple-Input Multiple-Output radar, through the wall radar, and statistical signal processing.
\end{IEEEbiography} 
\vspace{-28pt}
\begin{IEEEbiography}  [{\includegraphics[width=1in,height=1.25in,clip,keepaspectratio]{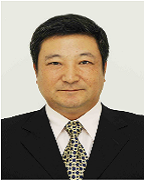}}] 
	{Xiaobo Yang} received the B.S. and M.S. degrees in electronic engineering from the University of Electronic Science and Technology of China (UESTC), Chengdu, China, in 1986 and 1988, respectively. He is currently a Professor with the School of Information and Communication Engineering, UESTC. His research interests include Multiple-input Multiple-output radar, through-the-wall radar, statistical signal processing, and intelligent signal processing.
\end{IEEEbiography} 

\end{document}